\documentclass[journal]{IEEEtran}

\usepackage{cite}
\usepackage{amsmath,amssymb,amsfonts}
\usepackage{algorithmic}
\usepackage{graphicx}
\usepackage{textcomp}
\usepackage{xcolor}
\usepackage{booktabs}
\def\BibTeX{{\rm B\kern-.05em{\sc i\kern-.025em b}\kern-.08em
    T\kern-.1667em\lower.7ex\hbox{E}\kern-.125emX}}
\usepackage{balance}
\usepackage{pifont}
\usepackage[T1]{fontenc}
\usepackage{multirow}
\usepackage{url}
\usepackage{balance}

\def\W{{\mathbf W}}
\def\x{{\mathbf x}}
\def\ii{{\hat{\imath}}}												
\def\bH{{\mathbb H}}
\def\bC{{\mathbb C}}
\def\bQ{{\mathbb Q}}

\begin{document}

\title{Multi-View Hypercomplex Learning \\for Breast Cancer Screening}

\author{Eleonora~Lopez, Eleonora~Grassucci, and~Danilo Comminiello,
\thanks{Authors are with the Department of Information Engineering, Electronics and Telecommunications (DIET), Sapienza University of Rome, Italy. \\Corresponding author's email: eleonora.lopez@uniroma1.it.}}

\maketitle

\begin{abstract}
Radiologists interpret mammography exams by jointly analyzing all four views, as correlations among them are crucial for accurate diagnosis. Recent methods employ dedicated fusion blocks to capture such dependencies, but these are often hindered by view dominance, training instability, and computational overhead. To address these challenges, we introduce \textit{multi-view hypercomplex learning}, a novel learning paradigm for multi-view breast cancer classification based on parameterized hypercomplex neural networks (PHNNs). Thanks to hypercomplex algebra, our models intrinsically capture both intra- and inter-view relations. We propose PHResNets for two-view exams and two complementary four-view architectures: PHYBOnet, optimized for efficiency, and PHYSEnet, optimized for accuracy. Extensive experiments demonstrate that our approach consistently outperforms state-of-the-art multi-view models, while also generalizing across radiographic modalities and tasks such as disease classification from chest X-rays and multimodal brain tumor segmentation. Full code and pretrained models are available at \url{https://github.com/ispamm/PHBreast}.
\end{abstract}


\IEEEpeerreviewmaketitle

\section{Introduction}
\label{sec:intro}

\begin{figure}[t]
    \centering
    \includegraphics[width=\linewidth]{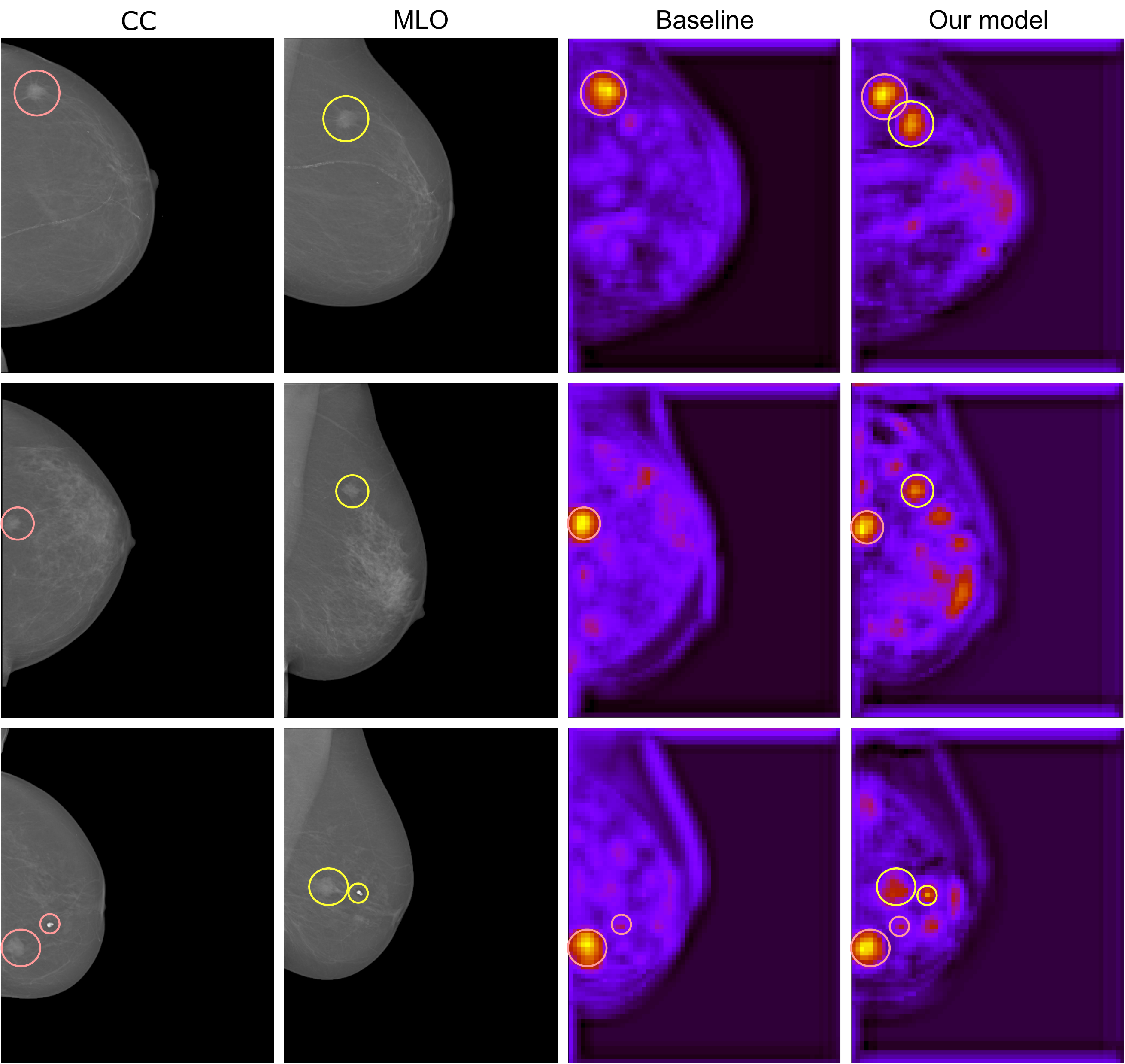}
    \caption{Visualization of hypercomplex multi-view learning in breast cancer classification. Malignant masses are highlighted in red and yellow in the respective views (CC and MLO). We compare activation maps of a real-valued baseline (SEnet) and our hypercomplex model (PHYSEnet). The proposed method learns from both views as shown by the highlighted areas corresponding to both views, while the conventional model focuses only on one view, discarding the information coming from the second one. 
    }
    \label{fig:activations}
\end{figure}

Among the different types of cancer that affect women worldwide, breast cancer alone accounts for almost one-third, making it by far the cancer with highest incidence among women \cite{Siegel2022Cancer}. For this reason, early detection of this disease is of extreme importance, and to achieve this, screening mammography is performed annually on all women above a certain age in regions where such technique is accessible \cite{MOREIRA2012236, Misra2010Screening}. During a mammography exam, two views of the breast are taken, capturing it from above, i.e., the craniocaudal (CC) view, and from the side, i.e., the mediolateral oblique (MLO) view. The CC and MLO views of the same breast are referred to as \textit{ipsilateral views}, while the same view (either CC or MLO) from both breasts are referred to as \textit{bilateral views}. When reading a mammogram, radiologists examine the views by performing a double analysis, comparing both ipsilateral views and bilateral views, as each perspective provides valuable diagnostic information. Such multi-view analysis has been found to be essential to make an accurate diagnosis of breast cancer \cite{Gur2009DBT, Liu2021Act}.

Recently, many works have been employing deep learning (DL)-based methods in the medical field and, especially, for breast cancer classification and detection with encouraging results \cite{zhou2025adaptive, alhussen2025xai, liang2024hradnet, li2025interpretable, shen2019Mammo, mo2023hover, iqbal2023bts, Wu2021ReducingFB, mahmood2024harnessing, ren2023ipsilateral, sreekala2025enhancing, wang2022wdccnet, shen2021interpretable, lopez2024attention, prinzi2024breast}. Inspired by the multi-view analysis performed by radiologists, several recent studies adopt a multi-view architecture to obtain a more robust and performing model \cite{Wu2020Screen, Wu2020MultViews, Zhang2020NewConv, Kyono2018MAMMOAD, Kyono2021Triage, kyono2019multi, Sun2019MVConv, YANG2021MommiNet, wu2024local, han2024deep}. A common strategy is to design separate branches for different views and later fuse their features, while more advanced approaches employ cross-attention or other dedicated fusion modules \cite{nakach2024comprehensive}. However, these methods can be difficult to train reliably, tend to over-rely on dominant views, or may compromise view-specific detail \cite{nakach2024comprehensive, Wu2020MultViews}. In some cases, the model might fail entirely in leveraging the correlated views and even worsen performance with respect to its single-view counterpart \cite{Wu2020MultViews}. 

In recent years, quaternion neural networks (QNNs) have gained a lot of interest in a variety of applications \cite{GaudetIJCNN2018, sigillo2025quaternion, ParcolletAIR2019, ParcolletICLR2019, ComminielloICASSP2019a, Brignone2022ISCAS, grassucci2023grouse}. Thanks to quaternion algebra rules, QNNs are effective in modeling relations within multidimensional data, unlike real-valued models \cite{ParcolletAIR2019, comminiello2024demystifying}. Furthermore, built upon the idea of QNNs, parameterized hypercomplex neural networks (PHNNs) generalize hypercomplex multiplications as a sum of $n$ Kronecker products, where $n$ is a hyperparameter that controls the domain in which the model operates \cite{Zhang2021PHM, grassucci2021lightweight}. They improve previous shortcomings by making these models applicable to any $n$-dimensional input (instead of just $3$D/$4$D as the quaternion domain) thanks to the introduction of the parameterized hypercomplex multiplication (PHM) and convolutional (PHC) layers \cite{Zhang2021PHM, grassucci2021lightweight}.

Motivated by the limitations of current multi-view approaches and the benefits of hypercomplex algebra, we propose \textit{multi-view hypercomplex learning}, a novel learning paradigm for breast cancer classification. Our approach rethinks how correlations across mammographic views are modeled, introducing a framework that captures both intra-view (global) and inter-view (local) dependencies, mimicking the radiologist's reading, as can be seen in \ref{fig:activations}. While the inductive bias of hypercomplex models is well established in multi-dimensional signals such as audio or RGB images \cite{comminiello2024demystifying, ParcolletICASSP2019a, grassucci2021lightweight}, its applicability to the structurally distinct multi-view domain has not been previously studied. To the best of our knowledge, in this work, we are the first to investigate and validate that this inductive bias indeed holds in multi-view medical imaging, where inter-view correlations play a crucial diagnostic role. To implement this paradigm, we design dedicated architectures that apply hypercomplex processing across views: PHResNets for two-view exams, and two complementary four-view models, PHYBOnet, optimized for efficiency, and PHYSEnet, optimized for accuracy, as detailed in Section~\ref{sec:method}.

We evaluate the effectiveness of our approach on two widely adopted benchmark datasets of mammography images, namely CBIS-DDSM \cite{Lee2017CBISdataset} andINbreast \cite{MOREIRA2012236}. Through extensive experiments, we demonstrate that our models surpass real-valued baselines and state-of-the-art methods. Finally, we further validate the proposed method on two different medical problems and radiographic techniques, i.e., CheXpert \cite{Chexpert2019Irvin} for chest X-rays and BraTS19 \cite{Brats2015Menze, brats2017bakas} for brain MRI, to prove the generalizability of the proposed approach.

The rest of the paper is organized as follows. Section~\ref{sec:problem} gives a detailed overview of the multi-view approach for breast cancer analysis. Section~\ref{sec:qnn} provides theoretical background on hypercomplex models, and Section~\ref{sec:method} presents the proposed method. The experiments are set up in Section~\ref{sec:exp_setup} and evaluated in Section~\ref{sec:exp_eval}. Finally, conclusions are drawn in Section~\ref{sec:conc}.

\section{Multi-View Approach in Breast Cancer Analysis}
\label{sec:problem}

\begin{figure}[t]
    \centering
    \includegraphics[width=0.8\linewidth]{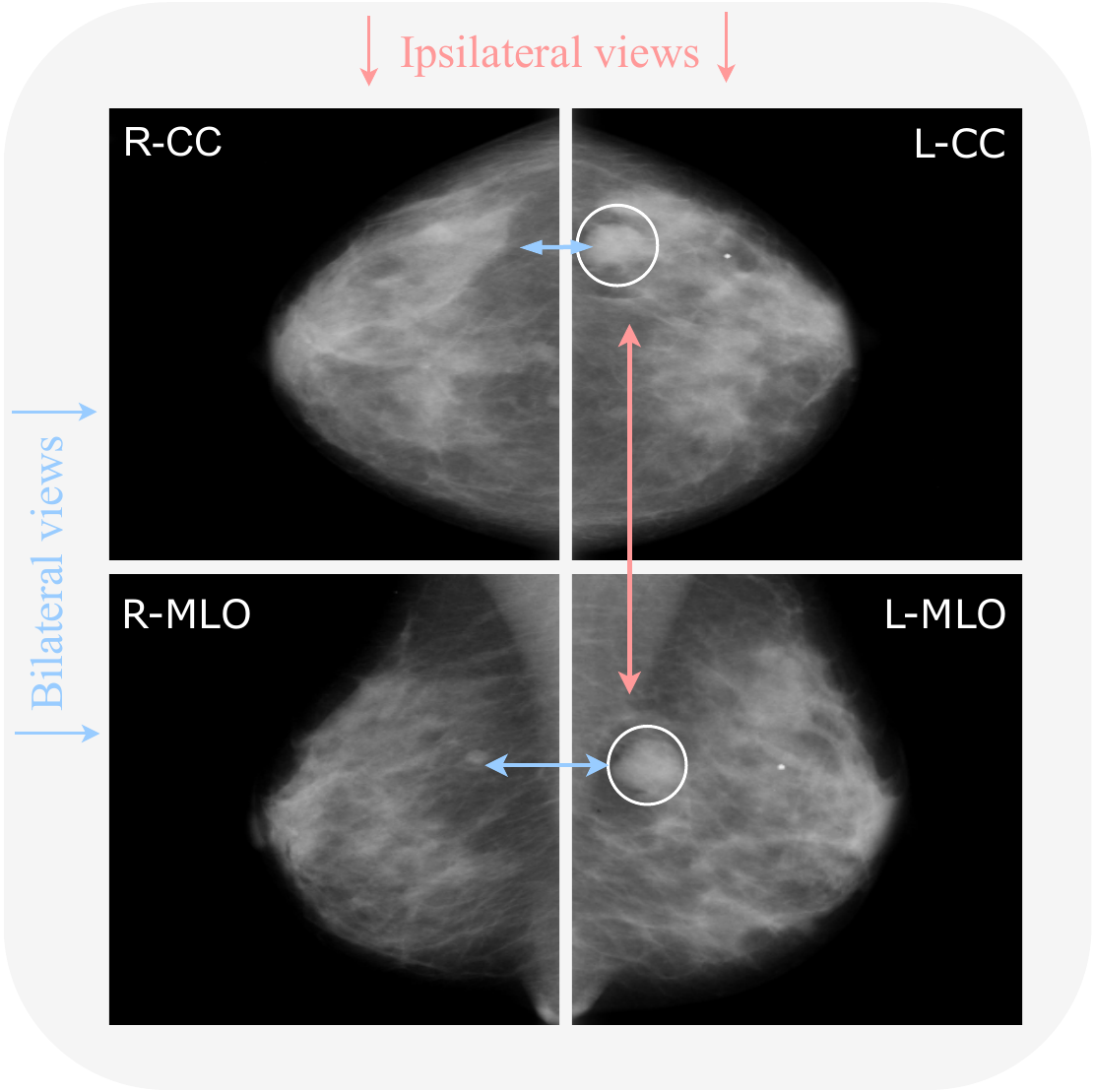}
    \caption{Example of a mammography exam. Images are from the CBIS-DDSM dataset and present an exam with four views: left CC (L-CC), right CC (R-CC), left MLO (L-MLO), and right MLO (R-MLO). Horizontal couples are bilateral views (blue), while vertical couples are ipsilateral views (red). Arrows indicate the relations among views. Bilateral relations reveal an asymmetry (as there is a mass on the left that is not present on the right). Ipsilateral relations give a better view of the specific mass.}
    \label{fig:cbis_example}
\end{figure}

Among the several imaging modalities, mammography is considered the best method for breast cancer screening and the most effective for early detection \cite{Misra2010Screening}. A mammography exam comprises four X-ray images produced by the recording of two views for each breast. The CC view is a top to bottom view, while the MLO view is a side view. During the reading of a mammography exam, employing multiple views is crucial in order to make an accurate diagnosis, as they contain complementary information. Admittedly, comparing ipsilateral views (CC and MLO views of the same breast) helps to detect eventual tumors and helps to analyze the 3D structure of masses \cite{Liu2021Act}. Whereas, studying bilateral views (same view of both breasts) helps to locate masses as asymmetries between them are an indicating factor \cite{Liu2021Act}. An example of a complete mammogram exam with ipsilateral and bilateral views is shown in Fig.~\ref{fig:cbis_example}.

Classic single-view models have been largely explored in the literature \cite{liang2024hradnet, shen2019Mammo, mo2023hover, Wu2021ReducingFB, ren2023ipsilateral, wang2022wdccnet, shen2021interpretable, lopez2024attention}. Among single-view approaches, a recent work has applied hypercomplex neural networks to attention-guided classification of mammograms \cite{lopez2024attention}. While effective in enhancing intra-view modeling, it does not address the multi-view setting. In contrast, our work is the first to extend hypercomplex learning to a multi-view scenario, explicitly modeling both intra- and cross-view dependencies and proposing dedicated architectures for both two-view and four-view exams.

Given the reading strategy employed by radiologists, many recent works have shifted toward a multi-view approach.
Indeed, studies have shown that a model employing multiple views can learn to generalize better compared to its single-view counterpart, while also reducing the number of false positives and false negatives \cite{shen2019Mammo, Wu2020Screen}.
Several works have leveraged multiple views using simple strategies, such as averaging predictions from two ipsilateral views \cite{shen2019Mammo, nakach2024comprehensive}.
Instead, more complex approaches involve designing multi-view architectures that leverage correlations during learning and not just at inference time. One of the most common architectural designs consists of multiple convolutional neural network (CNN) branches, each processing a different view \cite{nakach2024comprehensive, thakur2024systematic, nassif2022breast}. The outputs from these branches are typically concatenated and passed through fully connected layers to produce the final prediction  \cite{Wu2020MultViews, Wu2020Screen, Zhang2020NewConv, Kyono2018MAMMOAD, kyono2019multi, Kyono2021Triage, Sun2019MVConv}. One work takes this approach further by introducing cross-view bipartite graph-based reasoning following the CNN branches \cite{Liu2021Act, liu2020cross}. Another common strategy involves fusing information from different views using attention mechanisms \cite{nakach2024comprehensive, thakur2024systematic, nassif2022breast}. This includes integrating a cross-view attention module within a CNN \cite{zhao2020cross}, designing attention-based loss components \cite{chen2022multi}, employing multi-view vanilla transformers with various levels of fusion \cite{van2021multi, sun2022transformer}, or adopting multi-view Swin Transformer-based architectures \cite{sarker2024mvswint}.
The majority of these works focus on a two-view approach \cite{Wu2020MultViews, Zhang2020NewConv, Kyono2018MAMMOAD, Kyono2021Triage, Sun2019MVConv, chen2022multi, van2021multi, sarker2024mvswint, Liu2021Act, liu2020cross}, while some leverage the full four-view exam \cite{Wu2020Screen, kyono2019multi, zhao2020cross, sun2022transformer}.

Nonetheless, important challenges remain in how neural models leverage the information contained in multiple views. A study showed that commonly used multi-branch architectures can sometimes fail to exploit cross-view correlations effectively, resulting in the counter-intuitive outcome where a single-view model outperforms its multi-view counterpart \cite{Wu2020MultViews}. Attention-based fusion strategies and the addition of reasoning or loss components have shown promise, but they often suffer from training instability, are sensitive to imbalances where dominant views overshadow weaker ones, and can be computationally expensive \cite{nakach2024comprehensive}. Finally, transformer-based solutions face well-known limitations when applied to high-resolution images such as mammograms \cite{xu2025understanding}.

\section{Quaternion and Hypercomplex Neural Networks}
\label{sec:qnn}
Hypercomplex neural networks have recently attracted considerable attention due to their applicability across diverse domains, including images \cite{grassucci2022hypercomplex, bojesomo2024deep}, speech \cite{singh2024ux, panagos2024visual}, physiological signals \cite{lopez2023hypercomplex, lopez2024hierarchical, basso2023efficient, lopez2024phemonet}, and medical data \cite{yang2024parameterized, sigillo2025generalizing}.
Quaternion and hypercomplex neural networks are based on the hypercomplex number system $\bH$, which follows specific algebraic rules for addition and multiplication. Hypercomplex numbers generalize a plethora of algebraic systems, including complex numbers $\bC$, quaternions $\bQ$, and octonions $\mathbb{O}$, among others.
A generic hypercomplex number is defined as

\begin{equation}
    h = h_0 + h_1 \ii_1 + \ldots +  h_i \ii_i + \ldots + h_{n-1} \ii_{n-1},
\label{eq:hyp_num}
\end{equation}

\noindent whereby $h_0, \ldots, h_{n-1}$ are the real-valued coefficients and $\ii_1, \ldots, \ii_{n-1}$ the imaginary units. The first coefficient $h_0$ represents the real component, while the remaining ones compose the imaginary part. Therefore, the algebraic subsets of $\bH$ are identified by the number of imaginary units and by the algebraic rules that govern the interactions among them. 
Being part of the Cayley–Dickson family, hypercomplex algebras are defined at dimensions that are powers of two, i.e., $n=2^k$. Therefore, subset domains arise at predefined dimensions, i.e., $n=2,4,8,16, \ldots$. For other values of $n$, such as $n=3$, there is no canonical or widely accepted definition of multiplication, making them unsuitable for neural network design.
Moreover, with the increasing of $n$, Cayley-Dickson algebras lose some properties regarding vector multiplication, and more specific formulas need to be introduced to model this operation because of imaginary units interplays \cite{valle2021hypercomplex}. Indeed, as an example, quaternions and octonions products are not commutative since $\ii_1 \ii_2 \neq \ii_2 \ii_1$. Therefore, quaternion convolutional layers are based on the Hamilton product, according to which the convolutional filters $\W_0, \W_1, \W_2, \W_3$ are organized into a quaternion of the form $\W = \W_0 + \W_1 \ii_1 + \W_2 \ii_2 + \W_3 \ii_3$. The convolution operation between $\W$ and the quaternion input $\x = \x_0 + \x_1\ii_1 + \x_2\ii_2 + \x_3\ii_3$ can be then defined as:

\begin{equation}
{\bf{W}} * {\bf{x}} = \left[ {\begin{array}{*{20}c}
   \hfill {{\bf{W}}_0 } & \hfill { - {\bf{W}}_1 } & \hfill { - {\bf{W}}_2 } & \hfill { - {\bf{W}}_3 } \\
   \hfill {{\bf{W}}_1 } & \hfill {{\bf{W}}_0 } & \hfill { - {\bf{W}}_3 } & \hfill {{\bf{W}}_2 } \\
   \hfill {{\bf{W}}_2 } & \hfill {{\bf{W}}_3 } & \hfill {{\bf{W}}_0 } & \hfill { - {\bf{W}}_1 } \\
   \hfill {{\bf{W}}_3 } & \hfill { - {\bf{W}}_2 } & \hfill {{\bf{W}}_1 } & \hfill {{\bf{W}}_0 } \\
\end{array}} \right] * \left[ {\begin{array}{*{20}c}
   {{\bf{x}}_0 } \hfill  \\
   {{\bf{x}}_1 } \hfill  \\
   {{\bf{x}}_2 } \hfill  \\
   {{\bf{x}}_3 } \hfill  \\
\end{array}} \right],
\label{eq:qconv}
\end{equation}

\noindent in which the weight matrix on the left is the quaternion $\W$ expressed in matrix form and with signs adjusted to perform the Hamilton product \cite{GrassucciQGAN2021}.

\begin{figure}[t]
    \centering
    \includegraphics[width=0.45\linewidth]{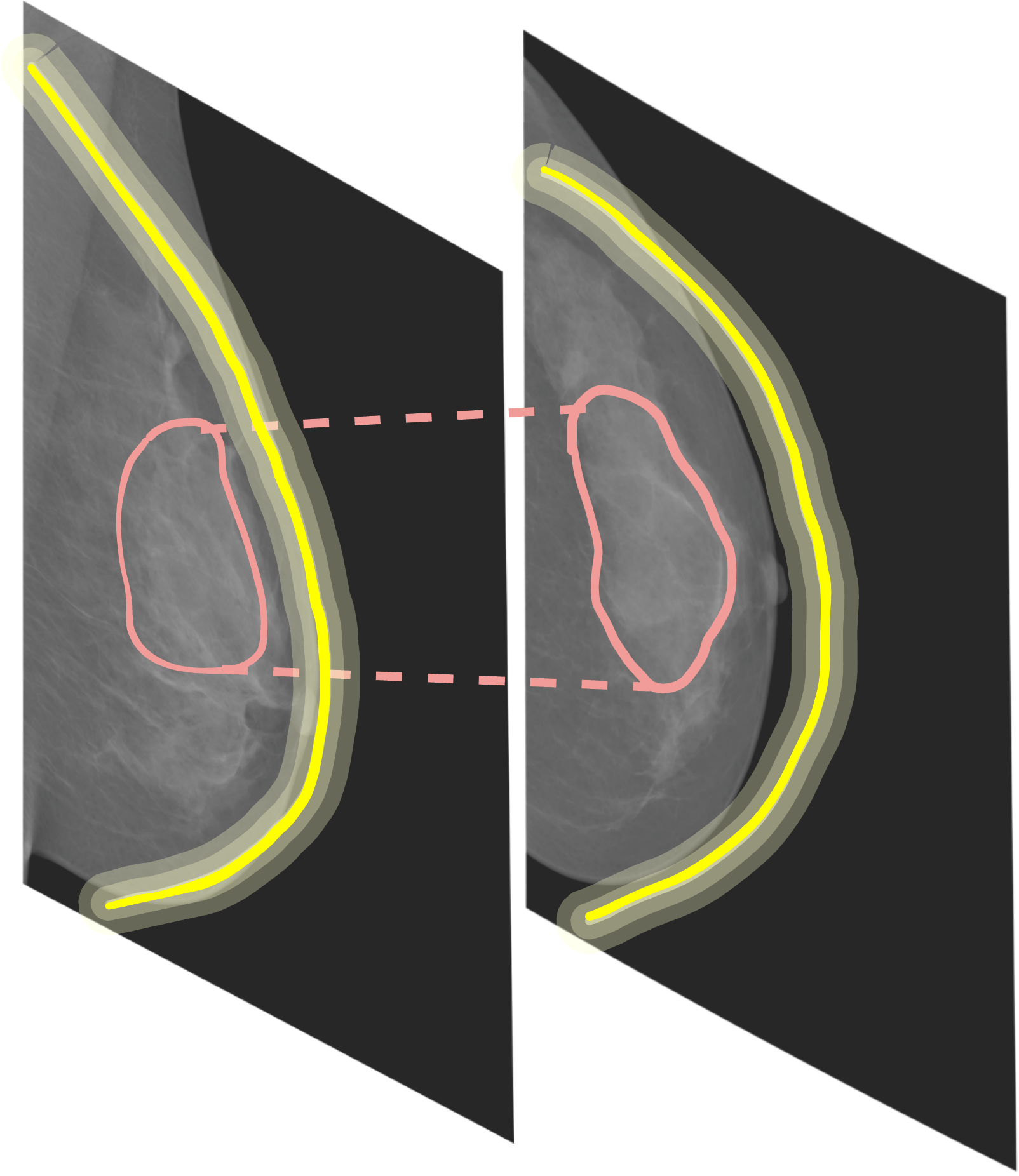}
    \caption{Global and local relations in multi-view mammography. Global relations refer to intra-view features, such as the overall breast shape within a single view. These are illustrated using different shades of yellow to represent variations in shape. Local relations refer to inter-view dependencies, such as shared or complementary textures across views, depicted in red.}
    \label{fig:global-local}
\end{figure}

Processing multidimensional inputs with quaternion convolutional neural networks (QCNNs) has several advantages. Indeed, due to the reusing of filter submatrices $\W_i, \; i=0,\ldots,3$ in eq.~\eqref{eq:qconv}, QCNNs are defined with $1/4$ free parameters with respect to real-valued counterparts with the same architecture structure. Moreover, sharing the filter submatrices among input components allows QCNNs to capture internal relations in input dimensions and to preserve correlations among them \cite{ParcolletAIR2019, comminiello2024demystifying}.
However, this approach is limited to $4$D inputs, thus various knacks are usually employed to apply QCNNs to different $3$D inputs, such as RGB color images. Recently, novel approaches proposed to parameterize hypercomplex multiplications and convolutions to maintain QCNNs and hypercomplex algebras advantages, while extending their applicability to any $n$D input \cite{Zhang2021PHM, grassucci2021lightweight}. The core idea of PHNNs is to develop the filter matrix $\W$ as a parameterized sum of Kronecker products:

\begin{equation}
    \W = \sum_{i=0}^n \mathbf{A}_i \otimes \mathbf{F}_i,
\label{eq:phc}
\end{equation}

\noindent whereby $n$ is a user-defined hyperparameter that determines the domain in which the model operates (i.e., $n=4$ for the quaternion domain, $n=8$ for octonions, and so on). The matrices $\mathbf{A}_i$ encode the algebra rules, that is the filter organization for convolutional layers, while the matrices $\mathbf{F}_i$ enclose the weight filters. Both these elements are completely learned from data during training, thus PHNNs can work also for values of $n$ for which an algebra does not yet exist. 
Indeed, PHNNs can process color images in their natural domain ($n=3$) without adding any uninformative channel (as previously done for QCNNs), while still exploiting latent relations between channels. Furthermore, PHC layers employ $1/n$ free parameters with respect to real-valued counterparts, so the user can govern both the domain and the parameters reduction by simply setting the value of $n$.

The reason behind the success of hypercomplex models is their ability to model not only global relations, as standard neural networks, but also local relations \cite{ParcolletAIR2019}. Global dependencies are related to spatial relations among pixels within the same channel (e.g., in an RGB image, global relations refer to patterns within a single R, G, or B channel). Instead, local dependencies implicitly link different channels. Neural models in the real domain tend to ignore local relations as they consider the pixels in different channels as sets of decorrelated points \cite{ParcolletAIR2019, comminiello2024demystifying}. In contrast, in a hypercomplex network, global relations are captured through standard convolutional operations (as in real-valued networks), while local relations are modeled via the structured parameter sharing rooted in the algebraic formulation of hypercomplex convolution (see Eq.~\ref{eq:qconv}). This property introduces an inductive bias that encourages the learning of correlations across input dimensions, i.e., local relations, and therefore allows hypercomplex models to simultaneously capture both global and local dependencies.

\section{Proposed Method}
\label{sec:method}

\begin{figure}[t]
    \centering
    \includegraphics[width=\linewidth]{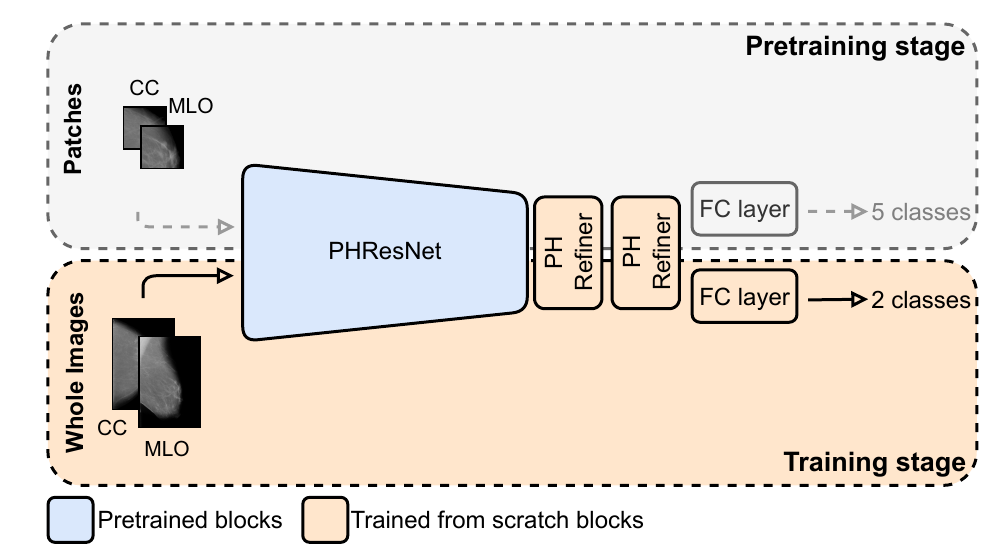}
    \caption{Training pipeline and two-view model overview. Blocks in blue, indicate pretrained blocks, while blocks in orange indicate blocks trained from scratch. The top part of the figure corresponds to pretraining phase, and the bottom part to training on whole images. During pretraining, the PHResNet backbone is trained on patches extracted from the original mammogram. They are classified into five classes: background or normal, benign and malignant calcifications, and masses, respectively. Then, when training on whole images, the PHResNet backbone is initialized with patch classifier weights and two PH convolutional refiner blocks are added and trained from scratch together with the final classification layer. Whole images are classified into 2 classes: malignant or benign/normal.}
    \label{fig:pipeline}
\end{figure}

\begin{figure*}[t]
    \centering
    \includegraphics[width=0.75\textwidth]{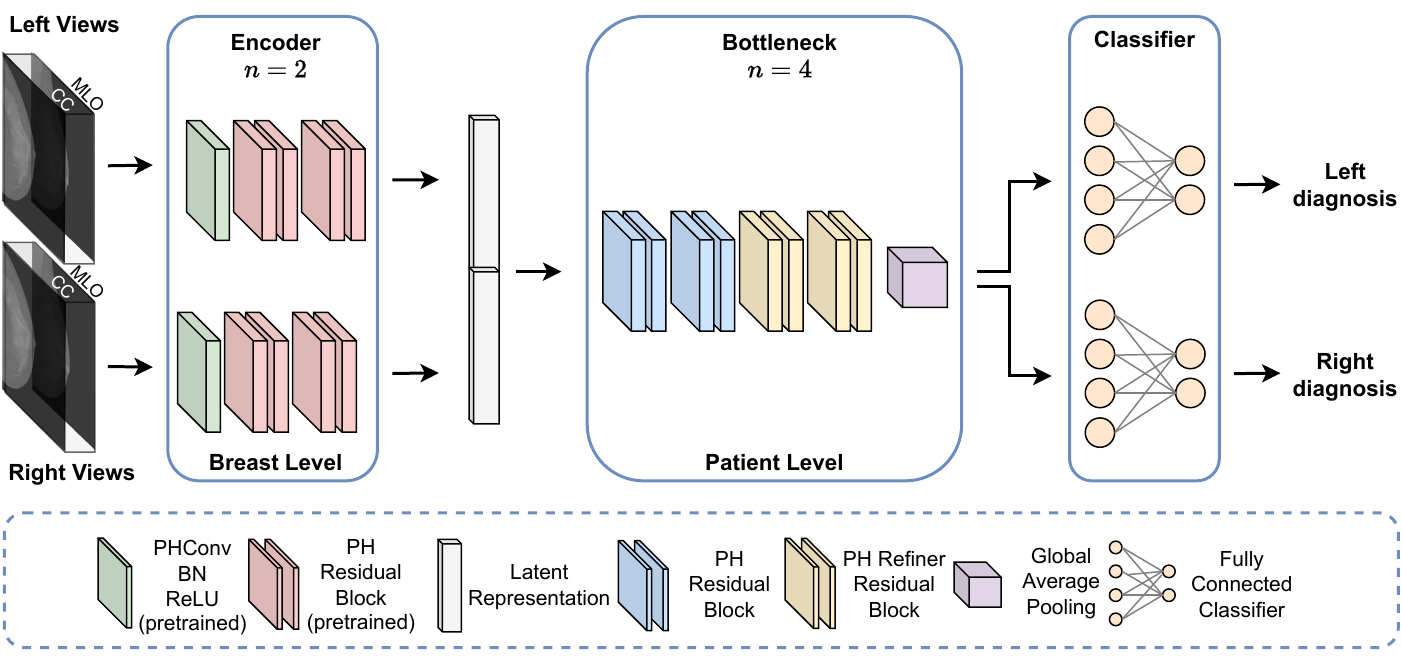}
    \caption{PHYBOnet architecture. A lightweight four-view model that performs breast-level classification by encoding each breast ipsilateral views through dedicated hypercomplex encoders defined with $n=2$. These encoders produce latent representations that are fused in a central bottleneck module defined with $n=4$ for a patient-level analysis.}
    \label{fig:fourview18}
\end{figure*}

\subsection{Multi-view Hypercomplex Learning}
We define a novel learning paradigm, i.e., \textit{multi-view hypercomplex learning}. While global and local relations have been defined in the context of multi-dimensional data (as discussed in Section~\ref{sec:qnn}), these notions differ when considering multi-view inputs. Therefore, we redefine global and local relations in the context of multi-view data, e.g., a mammography exam. Global relations are defined as intra-view features that capture patterns within a single mammogram view. Instead, local relations are formulated as inter-view features, i.e., the multi-view information contained in different mammogram views, as depicted in Fig.~\ref{fig:global-local}.
Although hypercomplex models are known to encode local relations across input dimensions, it remained unclear whether this property would also hold in the multi-view domain. 
We show that, thanks to their inductive bias, hypercomplex models inherently encode multi-view features as local dependencies, which are typically ignored by real-valued networks \cite{ParcolletAIR2019}. We refer to this process as \textit{multi-view hypercomplex learning}.

\subsection{Two-view hypercomplex model}
\label{subsec:arch2views}

The proposed multi-view architecture in the case of two views is a PHResNet set with the hyperparameter $n=2$ with additional hypercomplex refiner blocks. The complete model is depicted in Fig.~\ref{fig:pipeline} together with the training strategy. For simplicity, we refer to it as PHResNet. 

A ResNet block is typically defined by:

\begin{equation}
    \mathbf{y} = \mathcal{F}(\mathbf{x}) + \mathbf{x},
\label{eq:resnet}
\end{equation}

\noindent where $\mathcal{F}(\x)$ is composed by interleaving convolutional layers, batch normalization (BN) and ReLU activation functions. When equipped with PHC layers, real-valued convolutions are replaced with PHC to build PHResNets, therefore $\mathcal{F}(\x)$ becomes:

\begin{equation}
    \mathcal{F}(\mathbf{x}) = \text{BN} \left(\text{PHC} \left(\text{ReLU} \left(\text{BN} \left(\text{PHC} \left(\mathbf{x} \right)\right)\right)\right)\right),
\end{equation}

\noindent in which $\mathbf{x}$ is the multi-view input. In detail, the input of the model consists of the ipsilateral views, which are fed to the network as a multidimensional input (channel-wise). 
They are processed by a PHResNet backbone and then by two hypercomplex refiner blocks. The latter are additional hypercomplex residual blocks designed to refine the information learned by the PHResNet backbone, similarly to \cite{shen2019Mammo}. Finally, the network outputs a binary prediction indicating the presence of either a malignant or benign/normal finding (depending on the dataset).

\subsection{Four-view hypercomplex models}
\label{subsec:arch4views}

In this section, we describe the two four-view architectures we propose, i.e., PHYBOnet, a lightweight yet competitive model, and PHYSEnet, a more performant architecture.

\subsubsection{PHYBOnet}

The first model we propose is shown in Fig.~\ref{fig:fourview18}, namely the Parameterized Hypercomplex Bottleneck network (PHYBOnet). PHYBOnet is a lightweight alternative that yields competitive results. Initially, it focuses on the breast level and subsequently on the patient level. It consists of two encoder branches for each breast side with $n=2$, a bottleneck with the hyperparameter $n=4$, and two final classifier layers that produce the binary prediction for each corresponding side. Each encoder takes as input the ipsilateral views and has the objective of learning a latent representation. The learned embeddings are then merged and processed by the bottleneck with $n=4$, applying multi-view hypercomplex learning at the embedding level between the four view embeddings. It is a light network as the number of parameters is reduced by $1/4$.

\subsubsection{PHYSEnet}

\begin{figure*}[t]
    \centering
    \includegraphics[width=0.8\textwidth]{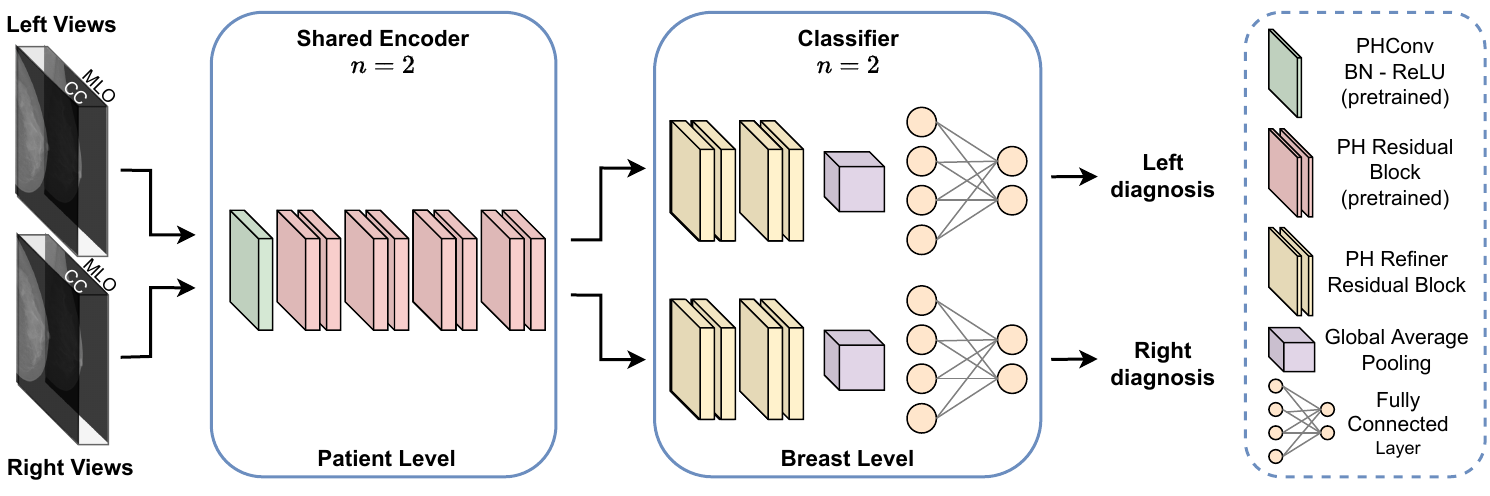}
    \caption{PHYSENet architecture. A four-view model that begins with a shared hypercomplex encoder defined with $n=2$ to process both pairs of ipsilateral views, focusing on a patient-level analysis. The resulting latent representations are refined through dedicated hypercomplex classifier branches with $n=2$ for breast-level classification.}
    \label{fig:fourviewbwm}
\end{figure*}

The second architecture we present, namely the Parameterized Hypercomplex Shared Encoder network (PHYSEnet), is depicted in Fig.~\ref{fig:fourviewbwm}. It has a broader focus on the patient-level analysis through an entire PHResNet18 with $n=2$ as the encoder model, which takes as input two ipsilateral views.  The weights of the encoder are shared between left and right inputs to jointly analyze the whole information of the patient. Then, two final classification branches, consisting of hypercomplex refiner blocks and a final fully connected layer, perform a breast-level analysis to output the final prediction for each breast. This design allows the model to leverage information from both ipsilateral and bilateral views. Indeed, thanks to multi-view hypercomplex learning, ipsilateral information is leveraged as in the case of two views, while by sharing the encoder between the two sides, the model is also able to take advantage of bilateral information.

\subsection{Pre-training procedure}
\label{subsec:training}

Training a classifier from scratch for this kind of task is very challenging for several reasons. First, neural models require large amounts of data for training, yet only a few publicly available breast cancer datasets exist, and they contain a limited number of examples. Additionally, a lesion occupies only a small portion of the original image, making it arduous to be detected by a neural network \cite{shen2019Mammo}.

To overcome these challenges, we deploy an \textit{ad hoc} pretraining strategy. We pretrain the model on patches of mammograms using only the PHResNet backbone, without refiner blocks, to perform five-class classification: background or normal, benign calcification, malignant calcification, benign mass, and malignant mass \cite{shen2019Mammo}. Patches are extracted by selecting 20 per lesion, 10 around the region of interest (ROI) and 10 from background or normal tissue. We consider two views also at the patch level, i.e., for all lesions that are visible in both views of the breast, patches around that lesion are taken for both views. Thus, the patch classifier takes as input two-view $224 \times 224$ patches of the original mammogram, concatenated along the channel dimension. This pretraining stage is illustrated in the top part of Fig.~\ref{fig:pipeline}. 
During whole-image training, the PHResNet is initialized with the pretrained patch classifier weights, and the final architecture is built as described in Section~\ref{subsec:arch2views}, including the additional hypercomplex refiner blocks and final classification layer, which are trained from scratch. The network is then fine-tuned for binary classification as malignant or benign/normal. The refiner blocks are designed to enhance the representations learned during pretraining and adapt them to whole-image reasoning. This stage is illustrated in the bottom part of Fig.~\ref{fig:pipeline}.  Pretraining in this way helps preserve fine-grained lesion information that might otherwise be lost during the downscaling required to process high-resolution mammograms (see Section~\ref{sec:data} for full-view preprocessing details) \cite{shen2019Mammo, Wu2020Screen}.

\section{Experimental setup}
\label{sec:exp_setup}

\subsection{Data}
\label{sec:data}

\subsubsection{CBIS-DDSM}
The Curated Breast Imaging Subset of DDSM (CBIS-DDSM) \cite{Lee2017CBISdataset} is a standardized version of the Digital Database for Screening Mammography (DDSM). It contains 2478 scanned film mammography images and it provides both pixel-level and whole-image labels from biopsy-proven pathology results and official training/test splits. Additionally, it provides separate splits for cases with masses and calcifications. CBIS-DDSM is employed for the training of the patch classifier as well as the whole-image classifier in the two-view scenario. It is not used for four-view experiments as the official training/test splits do not contain enough full-exam cases and creating different splits would result in data leakage between patches and whole images. The images are resized to $600 \times 500$ and are augmented with random rotations between $-25$ and $+25$ degrees and random horizontal and vertical flips \cite{shen2019Mammo}.

\subsubsection{INbreast}
INbreast \cite{MOREIRA2012236}, is a database of 410 full-field digital mammography (FFDM) images. The data splits are not provided, thus we manually create them by splitting the dataset patient-wise in a stratified fashion, using 20\% of the data for testing. Finally, INbreast does not provide pathological confirmation of malignancy but BI-RADS labels. Therefore, we consider BI-RADS categories 4, 5, and 6 as positive and 1, 2 as negative, whilst ruling out category 3 \cite{shen2019Mammo}. INbreast is utilized for experiments in both the two-view and four-view scenarios, and the same preprocessing as CBIS-DDSM is applied. 


\subsubsection{CheXpert}

The CheXpert \cite{Chexpert2019Irvin} dataset contains 224,316 chest X-rays with both frontal and lateral views and provides labels corresponding to 14 common chest radiographic observations. Images are resized to $320 \times 320$ \cite{Chexpert2019Irvin} and the same augmentation operations as before are applied. 

\subsubsection{BraTS19}
BraTS19 \cite{Brats2015Menze, brats2017bakas, brats2018bakas} is a dataset of multimodal brain MRI scans, providing four modalities for each exam, segmentation maps, demographic information (such as age) and overall survival (OS) data for 210 patients. Volumes are resized to $128 \times 128 \times 128$ and augmented via random scaling between 1 and 1.1 \cite{Post-hocOSCarneiro2021} for the task of OS prediction. Instead, for segmentation, 2D axial slices are extracted from the volumes, and the original shape of $240 \times 240$ is maintained.

\subsection{Evaluation metrics}
\label{subsec:metrics}

We adopt AUC (Area Under the ROC Curve) and accuracy to evaluate our models, as they are the most common metrics employed in medical imaging tasks \cite{Wu2020MultViews, Wu2020Screen, shen2019Mammo, Zhang2020NewConv}. We also report AUC-PR (Area Under the Precision-Recall Curve), which better reflects performance in imbalanced settings by focusing on the positive class. Sensitivity and specificity are measured at Youden’s index, the threshold that maximizes the sum of the two. Youden’s index provides an interpretable balance between true positive and true negative rates. Finally, we employ the Dice score for the segmentation task, which measures the pixel-wise agreement between a predicted mask and its corresponding ground truth.

\subsection{Implementation}
\label{subsec:implementation}

\subsubsection{Two-view architectures}
We perform the experimental validation with two ResNet backbones, i.e., PHResNet18 and PHResNet50. We omit the max pooling operation and we apply global average pooling, after which we add 4 hypercomplex refiner residual blocks constructed with the bottleneck design \cite{Resnet2016}. The backbone network is initialized with the patch classifier weights, while the refiner blocks and final layer are trained from scratch \cite{shen2019Mammo}.

\subsubsection{Four-view architectures}

\textbf{PHYBOnet.} The encoders comprise a first $3 \times 3$ convolutional layer with a stride of 1, together with batch normalization and ReLU, and 4 residual blocks, as shown in Fig.~\ref{fig:fourview18}. The bottleneck is composed of $8$ residual blocks and a global average pooling layer, with the last 4 blocks being the hypercomplex refiner blocks.

\textbf{PHYSEnet}. The encoder is a PHResNet18, while the two classifier branches are comprised of the 4 hypercomplex refiner blocks with a global average pooling layer and the final classification layer. 

\textbf{Weights initialization.} During training, both PHYBOnet and PHYSEnet have their encoder portions initialized with patch classifier weights, while the remaining parts of the networks are trained from scratch. In a second set of experiments conducted only with PHYSEnet, the whole architecture is initialized with the weights of the best whole-image two-view classifier trained on CBIS-DDSM.

\subsection{Training}
We train our models to minimize the binary cross-entropy loss using the Adam optimizer, with a learning rate of $10^{-5}$ for classification experiments and $2 \times 10^{-4}$ for segmentation. For two-view models, the batch size is 8 for PHResNet18, 2 for PHResNet50, and 32 for PHUNet. For four-view models, the batch size is set to 4. To handle imbalanced data in the four-view scenario, we use a weighted loss function, assigning a weight to positive examples equal to the ratio of benign/normal examples to positive examples. Instead, for the OS task we use the MSE loss. In each case, we adopt two regularization techniques, i.e., weight decay at $5 \times 10^{-4}$ and early stopping.

\begin{table}[t]
\centering
\caption{Results for two-view models without pretraining on CBIS-DDSM mass split (top) and combined mass + calcification split (bottom).}
\resizebox{\columnwidth}{!}{
\begin{tabular}{lccc}
\toprule
\multicolumn{1}{l}{Model} & \multicolumn{1}{c}{Params} & \multicolumn{1}{c}{AUC} & \multicolumn{1}{c}{Accuracy (\%)} \\ 
\midrule 
ResNet18 & 11M & 0.646 $\pm$ 0.008 & 64.554 $\pm$ 2.846  \\
ResNet50 & 16M & 0.663 $\pm$ 0.011 & 67.606 $\pm$ 1.408\\
MV-Swin-T \cite{sarker2024mvswint} & 29M & \textbf{0.714} $\pm$ ------- & \underline{68.630} $\pm$ ------- \\
PHResNet18 (ours) & 5M & 0.660 $\pm$ 0.020 & 67.371 $\pm$ 2.846 \\ 
PHResNet50 (ours) & 8M &\underline{0.700} $\pm$ 0.002 & \textbf{70.657} $\pm$ 1.466 \\ 
\midrule
ResNet18  & 11M & 0.644 $\pm$ 0.021 & 65.349 $\pm$ 0.823 \\
ResNet50 & 16M & 0.651 $\pm$ 0.017 & \textbf{66.271} $\pm$ 1.826\\
MV-Swin-T \cite{sarker2024mvswint} & 29M & \underline{0.664} $\pm$ ------- & 65.370 $\pm$ -------\\
PHResNet18 (ours) & 5M & 0.661 $\pm$ 0.009 & 64.954 $\pm$ 1.597 \\ 
PHResNet50 (ours) & 8M & \textbf{0.667} $\pm$ 0.011 & \underline{66.008} $\pm$ 1.581 \\ 
\bottomrule   
\end{tabular}
}
\label{tab:cbis_2views_no_pretr}
\end{table}

\begin{table}[t]
\centering
\caption{Results for patch classifiers on CBIS-DDSM dataset.}
\label{tab:patch}
\begin{tabular}{lcc}
\toprule
Model       & \multicolumn{1}{c}{Accuracy (\%)} & \multicolumn{1}{c}{Pretrained}\\ \midrule 
EfficientNet-B0 \cite{petrini2022breast} & 75.540 & \ding{51} \\ \midrule
ResNet18    & 74.942 & \multirow{2}{*}{\ding{55}} \\
PHResNet18 & \textbf{76.825} &  \\ \midrule
ResNet50    & 75.989 & \multirow{2}{*}{\ding{55}} \\
PHResNet50 & \textbf{77.338} & \\
\bottomrule   
\end{tabular}
\end{table}

\begin{table*}[t]
\centering
\caption{Results for two-view models with pretraining. For the CBIS-DDSM dataset, we perform experiments on the mass split (top) and the combined mass + calcification split (middle), using pretraining on patches. For the INbreast dataset, we pretrain the models on patches and then on whole CBIS images (Patches + CBIS). 
}
\resizebox{\linewidth}{!}{
\begin{tabular}{llccccc}
\toprule
\multicolumn{1}{l}{Dataset} & \multicolumn{1}{l}{Model} & \multicolumn{1}{c}{Params} &\multicolumn{1}{c}{Pretraining} & \multicolumn{1}{c}{AUC} & \multicolumn{1}{c}{Accuracy (\%)} \\ \midrule 
\multirow{11}{*}{CBIS (mass)} & ResNet18 & 26M & \multirow{11}{*}{Patches} & 0.710 $\pm$ 0.018 & 70.892 $\pm$ 3.614 \\
& ResNet50 & 32M & & 0.724 $\pm$ 0.007 & 73.474 $\pm$ 1.076 \\
& Shared ResNet \cite{Wu2020MultViews} & 12M & & 0.735 $\pm$ 0.014 & 72.769 $\pm$ 2.151 \\
& Breast-wise-model \cite{Wu2020Screen} & 23M & & 0.705 $\pm$ 0.011 & 69.484 $\pm$ 2.151 \\
& DualNet \cite{Rubin2018LargeSA} & 13M & & 0.705 $\pm$ 0.018 & 69.719 $\pm$ 1.863 \\
& Two-View EfficientNet-B0 \cite{petrini2022breast} & 86M & & 0.692 $\pm$ ------- & 69.718 $\pm$ ------- \\
& Cross-view transformer \cite{van2021multi} & 24M & & 0.724 $\pm$ 0.026 & 71.596 $\pm$ 1.772 \\
& BRAIxMVCCL \cite{chen2022multi} & 27M & & 0.672 $\pm$ 0.013 & 69.014 $\pm$ 1.220 \\
& MV-Swin-T \cite{sarker2024mvswint} & 30M & & 0.721 $\pm$ 0.002 & 70.789 $\pm$ 1.408 & \\
& PHResNet18 (ours) & 13M & & \underline{0.737} $\pm$ 0.004 & \underline{74.882} $\pm$ 1.466 \\
& PHResNet50 (ours) & 16M & & \textbf{0.739} $\pm$ 0.004 & \textbf{75.352} $\pm$ 1.409 \\ \midrule
\multirow{5}{*}{CBIS (mass and calc.)} & ResNet18  & 26M & \multirow{5}{*}{Patches} & 0.659 $\pm$ 0.012 & 66.271 $\pm$ 1.271 \\
& ResNet50 & 32M & & 0.659 $\pm$ 0.013 & 65.217 $\pm$ 3.236 \\
& Two-View EfficientNet-B0 \cite{petrini2022breast} & 86M& & 0.644 $\pm$ ------- & 65.613 $\pm$ ------- \\
& PHResNet18 (ours) & 13M & & \textbf{0.677} $\pm$ 0.005 & \textbf{68.116} $\pm$ 1.388 \\
& PHResNet50 (ours) & 16M & & \underline{0.676} $\pm$ 0.014 & \underline{67.062} $\pm$ 0.995 \\ \midrule
\multirow{4}{*}{INbreast} & ResNet18 & 26M & \multirow{4}{*}{Patches + CBIS} & \underline{0.789} $\pm$ 0.073 & \underline{81.887} $\pm$ 1.509 \\
& ResNet50 & 32M & & 0.755 $\pm$ 0.063 & 77.358 $\pm$ 7.825 \\
& PHResNet18 (ours) & 13M & & \textbf{0.793} $\pm$ 0.071 & \textbf{83.019} $\pm$ 5.723 \\
& PHResNet50 (ours) & 16M & & 0.759 $\pm$ 0.045 & 80.000 $\pm$ 6.382 \\
\bottomrule   
\end{tabular}
}
\label{tab:cbis_2views}
\end{table*}

\section{Experimental evaluation}
\label{sec:exp_eval}

In this section, we present an exhaustive evaluation of our method. For all experiments, we report the mean AUC and accuracy over multiple runs (three random runs unless otherwise specified) together with the standard deviation. In each table, the best and second-best results are highlighted in bold and underlined, respectively. These results demonstrate that, under fair comparative conditions, our method surpasses existing approaches by effectively utilizing multi-view information thanks to multi-view hypercomplex learning.

\subsection{Preliminary experiments}
\label{sec:prel}

\subsubsection{Whole-image without pretraining}
\label{sec:no_pretr}

We conduct preliminary experiments reported in Tab.~\ref{tab:cbis_2views_no_pretr} to evaluate the ability of the models to learn from data without any form of pretraining. We test five different architectures on whole mammograms of CBIS-DDSM. We compare PHResNet18 and PHResNet50 against their real-valued counterparts ResNet18 and ResNet50, and a recent method based on transformers \cite{sarker2024mvswint}. Herein, it is evident that all the models involved are not able to learn properly and struggle to discriminate between benign and malignant images, as expected and explained in Section~\ref{subsec:training}. Nevertheless, even with poor performance, it is already evident that the proposed models with multi-view hypercomplex learning are able to capture more information contained in the correlated views and thus reach a higher AUC and accuracy with respect to the real-valued counterparts in almost all the experiments. They also achieve results comparable to those of the more complex MV-Swin-T architecture \cite{sarker2024mvswint}.

\subsubsection{Patch classifier}

Preliminary experiments also include the pretraining phase of patch classifiers, which is carried out with the purpose of overcoming the problems described in Subsection~\ref{subsec:training}.
Table~\ref{tab:patch} reports the results of these experiments. We observe a significant performance gap at the patch level between our hypercomplex models and the real-valued ones. Moreover, our networks also surpass a method \cite{petrini2022breast} that includes pretraining on ImageNet, while ours are trained from scratch.

\subsection{Two-view experiments}
\label{sec:2views}

\subsubsection{Methods for comparison}
\label{sec:2views_comp}

We compare the proposed PHResNets with multi-view hypercomplex learning against the respective real-valued baseline models (ResNet18 and ResNet50) and seven state-of-the-art multi-view architectures \cite{Wu2020MultViews, Wu2020Screen, Rubin2018LargeSA, petrini2022breast, van2021multi, chen2022multi, sarker2024mvswint}. The model proposed in \cite{Rubin2018LargeSA} processes frontal and lateral chest X-rays, employing as backbone DenseNet121, while methods proposed in \cite{Wu2020MultViews, Wu2020Screen, petrini2022breast, van2021multi, chen2022multi, sarker2024mvswint} are designed for mammography images. To ensure a fair comparison with our own networks, they are all trained in the same experimental setup as our own models. For the Two-View EfficientNet-B0 network, we take the results directly from the paper as they already employ the same pretraining. Additionally, original architectures proposed in \cite{Wu2020MultViews, Wu2020Screen} employ as backbone a variation of the standard ResNet50, namely ResNet22, which was designed specifically for the purpose of handling high-resolution images. However, in our case the mammograms are resized as explained in Subsection~\ref{sec:data}, thus we use a standard ResNet18 instead. Finally, since \cite{Wu2020Screen} proposes networks designed to handle four views, to compare this approach with ours, we employ only the ResNet paths of the Breast-wise-model which handle two ipsilateral views.

\begin{figure}[t]
    \centering
    \includegraphics[width=\linewidth]{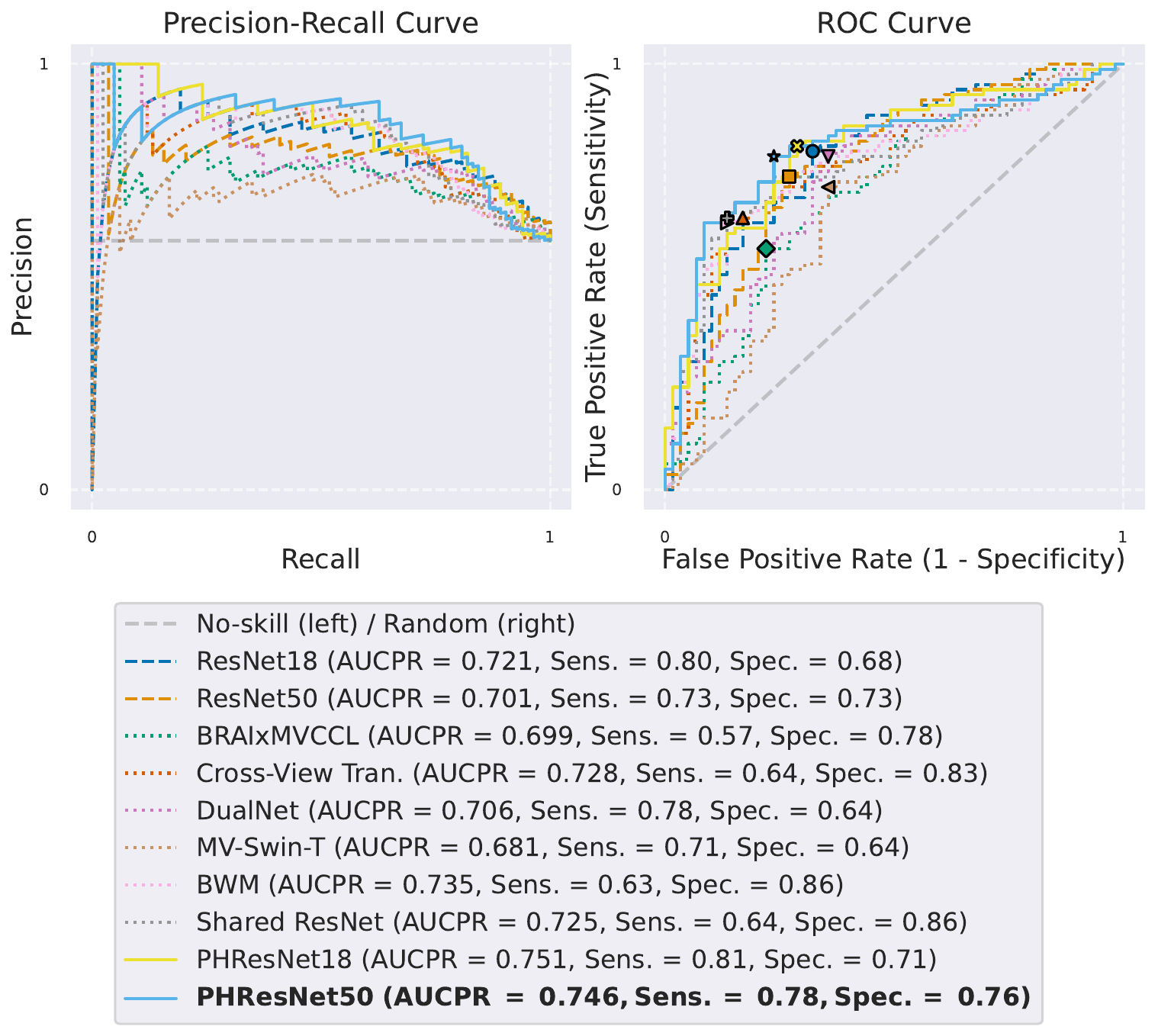}
    \caption{Precision-Recall and ROC curves for two-view models evaluated on the CBIS-DDSM mass subset. Points marked on the ROC curves indicate Youden’s index. In the legend, we report the AUCPR as well as the sensitivity and specificity at Youden’s index.}
    \label{fig:prauc}
\end{figure}

\subsubsection{Results}
\label{sec:2views_res}

The results of the experiments in the two-view scenario are reported in Tab.~\ref{tab:cbis_2views}, together with the number of parameters for each model. For CBIS-DDSM, models are pretrained on patches as described in Section~\ref{subsec:training}, while for INbreast, models are initialized with the weights of the best whole-image classifier trained on CBIS-DDSM. Firstly, the advantages of the employed pretraining strategy are clear by comparing the results with Tab.~\ref{tab:cbis_2views_no_pretr}. Most importantly, PHResNets clearly outperform both baseline counterparts implemented in the real domain and all other state-of-the-art methods. Despite being much smaller than the other methods, our hypercomplex models still outperform them by effectively leveraging correlations thanks to multi-view hypercomplex learning. Overall, PHResNet50 and PHResNet18 achieve similar results. This suggests that a larger model such as PHResNet50 is unnecessary for CBIS-DDSM and INbreast datasets, as PHResNet18 outperforms it in the mass and calcification split and INbreast experiments. 

Figure~\ref{fig:prauc} further supports the effectiveness of the proposed approach by showing the precision-recall and ROC curves of the models in the top part of Table~\ref{tab:cbis_2views}, with hypercomplex networks achieving the highest AUC-PR, sensitivity, and specificity. In conclusion, we have conducted statistical significance tests to further validate our results, specifically, two-sided paired \textit{t}-test. For instance, in the case of mass classification on CBIS-DDSM, the \textit{p}-value between PHResNet18 and ResNet18 is 0.057, while between PHResNet50 and ResNet50 is 0.022, which are both statistically significant for a significance level of 0.10.

\subsection{Four-view experiments}
\label{sec:4views}

\subsubsection{Methods for comparison}
\label{sec:4views_comp}

We compare the proposed architectures against the respective baseline implemented in the real domain (BOnet and SEnet), and further against four state-of-the-art approaches designed for breast cancer \cite{Wu2020Screen, zhao2020cross, sun2022transformer}. For \cite{Wu2020Screen}, we employ as comparison their best-performing model, the View-wise-model, along with the Breast-wise-model.
As in the two-view scenario, ResNet22 is replaced with ResNet18, and all models are trained in the same manner as ours to guarantee a fair comparison.

\begin{table*}[t]
\centering
\caption{Results for four-view models on INbreast. We pretrain models on CBIS-DDSM patches (top) as well as patches and then whole CBIS-DDSM images for further fine-tuning (bottom).
}

\begin{tabular}{lccc}
\toprule
Model & \multicolumn{1}{c}{Params} & \multicolumn{1}{c}{AUC} & \multicolumn{1}{c}{Accuracy (\%)} \\ \midrule 
BOnet & 27M & 0.756 $\pm$ 0.047 & 70.000 $\pm$ 4.216 \\
SEnet & 41M & \underline{0.786} $\pm$ 0.090 & 73.333 $\pm$ 12.824\\
View-wise-model \cite{Wu2020Screen} & 24M &  0.768 $\pm$ 0.072 & \underline{75.333} $\pm$ 6.182 \\
Breast-wise-model \cite{Wu2020Screen} & 24M &  0.734 $\pm$ 0.027 & 72.667 $\pm$ 7.717 \\
ResNet50+CvAM \cite{zhao2020cross} & 27M & 0.757 $\pm$ 0.029 & 75.109 $\pm$ 5.439 \\
Multi-view Swin-T \cite{sun2022transformer} & 37M & 0.772 $\pm$ 0.015 & 75.593 $\pm$ 4.102 \\
PHYBOnet (ours) & 7M &  0.764 $\pm$ 0.061 & 70.000 $\pm$ 7.601 \\
PHYSEnet (ours) & 20M & \textbf{0.798} $\pm$ 0.071 & \textbf{77.333} $\pm$ 7.717 \\ \midrule
SEnet & 41M & 0.796 $\pm$ 0.096 & 79.333 $\pm$ 8.273 \\
PHYSEnet (ours) & 20M & \textbf{0.814} $\pm$ 0.060 & \textbf{82.000} $\pm$ 6.864 \\
\bottomrule   
\end{tabular}
\label{tab:cbis_4views}
\end{table*}


\subsubsection{Results}
\label{sec:4views_res}

Table~\ref{tab:cbis_4views} reports the results of experiments with four views on INbreast. Given the small size of the INbreast dataset, we conduct cross-validation using 5 different splits. Our PHYSEnet outperforms all other methods, showing that simultaneous exploitation of ipsilateral and bilateral views yields a more effective classifier. In particular, both PHYSEnet and PHYBOnet surpass their counterparts in the real domain, achieving so with half/a quarter of parameters. Interestingly, our lightweight alternative, PHYBOnet with $n=4$, is also able to compete with the best models in the state-of-the-art, surpassing the Breast-wise-model and having overall comparable performance with only 7M parameters. Furthermore, we have performed statistical significance tests to corroborate our findings. For instance, the \textit{p}-value between PHYSEnet and the state-of-the-art Breast-wise-model is 0.076 which is statistically significant for a significance level of 0.10. We also test PHYSEnet and SEnet using the weights of the best whole-image two-view model trained on CBIS-DDSM as initialization. Both models benefit from the learned features on whole images, gaining a boost in performance. However, greater improvement is attained by the hypercomplex network, which achieves the best performance yet.


\subsection{Generalizing beyond mammograms}
\label{sec:additional_tasks}

\subsubsection{Multi-view multi-label chest X-ray classification}

The task consists of classifying the $14$ most common chest diseases from multi-view chest X-rays. The evaluation is then conducted on a subset of five labels \cite{Chexpert2019Irvin}. Since the exam comprises two views, we employ a PHResNet with $n=2$ and compare it against its real-valued counterpart. As shown in Tab.~\ref{tab:chest_classification}, PHResNet outperforms the real-valued ResNet. We additionally compare our approach with the model that won the CheXpert challenge, i.e., DeepAUC \cite{deepauc2021yuan}. Their model is a single-view architecture, and the best performance was achieved through an ensemble of networks. To ensure a fair comparison, we conduct a new run using a single model and compare it with our approach, i.e., PHDeepAUC with $n=2$, which is a multi-view adaptation of DeepAUC equipped with multi-view hypercomplex learning. Additionally, we modify the original DeepAUC to handle multiple views for a more straightforward comparison with our method. As shown in Tab.~\ref{tab:chest_classification}, PHDeepAUC surpasses the best model proposed in the original challenge. Moreover, it is noteworthy that the multi-view version of DeepAUC performs worse than the original single-view model. Indeed, as discussed in Section~\ref{sec:problem}, these types of networks tend to overfit on one of the two views, leading to the single-view model outperforming its multi-view counterpart \cite{Wu2020MultViews}. This result further validates the efficacy of our multi-view approach based on hypercomplex algebra, namely multi-view hypercomplex learning, which instead effectively exploits information from both views.

\subsubsection{Multimodal OS prediction}

\begin{table}[t]
\centering
\caption{Results for multi-view multi-label classification on CheXpert.}
\begin{tabular}{lcc}
\toprule
\multicolumn{1}{l}{Model} & \multicolumn{1}{c}{Params} & \multicolumn{1}{c}{AUC} \\ \midrule 
ResNet18 & 11M & 0.645 $\pm$ 0.164  \\
PHResNet18 & 5M &  0.722 $\pm$ 0.184 \\
DeepAUC (single-view) \cite{deepauc2021yuan} & 6M & \underline{0.834} $\pm$ 0.002 \\
DeepAUC (multi-view) & 6M & 0.820 $\pm$ 0.012 \\
PHDeepAUC (n=2) & 3M & 	\textbf{0.867 $\pm$ 0.009} \\
\bottomrule
\end{tabular}
\label{tab:chest_classification}
\end{table}

\begin{table}[t]
\centering
\caption{Results on BraTS19 for OS prediction (top) and brain tumor segmentation (bottom).}
\begin{tabular}{lcc}
\toprule
\multicolumn{1}{l}{Model} & \multicolumn{1}{c}{Params} & \multicolumn{1}{c}{Acc./Dice} \\ \midrule 
Post-hoc \cite{Post-hocOSCarneiro2021} & 310K & 0.414 \\
PH Post-hoc ($n=4)$ & 91K & \underline{0.448} \\
ResNet18 3D & 33M & 0.414 \\
PHResNet18 3D ($n=2$) & 16M & 	\textbf{0.517} \\
\midrule
UNet & 31M & \textbf{0.826} $\pm$ 0.017\\
PHUNet & 15M & 0.825 $\pm$ 0.018\\
\bottomrule
\end{tabular}
\label{tab:brats}
\end{table}

OS prediction is one of the official tasks of the BraTS19 challenge \cite{brats2018bakas} and consists of predicting the number of survival days. All models are evaluated on the provided validation set through the official online evaluation platform of the challenge. The task evaluation is conducted by considering three different classes: short-term (less than 10 months), mid-term (between 10 and 15 months), and long-term (more than 15 months) survivors. We consider two multi-view hypercomplex models, namely a simple PHResNet18 3D and a state-of-the-art method, Post-hoc \cite{Post-hocOSCarneiro2021}, in the hypercomplex domain, i.e., PH Post-hoc. The former takes as input two MRI modalities, T2 and FLAIR, so we set $n=2$, while the latter takes all four modalities of BraTS19 \cite{brats2018bakas} thus we set $n=4$. Moreover, both methods take in input the age parameter which is processed by a linear layer and a PHM layer with $n=1$, respectively. In both cases, the embeddings of the MRIs and age are summed together \cite{Post-hocOSCarneiro2021} and then processed by two fully connected layers to produce the final output. As we show in the top section of Tab.~\ref{tab:brats}, both hypercomplex networks surpass their respective real-valued counterparts. PHResNet18 3D achieves the highest accuracy, surpassing the state-of-the-art approach in both real and hypercomplex domains. Moreover, PH Post-hoc, which comprises only 91K parameters, outperforms ResNet18 3D, which instead has 33M parameters. This result further highlights the effectiveness of multi-view hypercomplex learning which allows to properly leverage both global and local relations, unlike real-valued networks.

\subsubsection{Multimodal brain tumor segmentation}

To demonstrate the versatility of the proposed approach, we also apply it to segmentation tasks. In particular, we conduct whole-tumor segmentation from multimodal slices extracted from MRI volumes. We define a UNet \cite{UNetRonneberger2015} in the hypercomplex domain, i.e. PHUNet, and set $n=2$ considering T2 and FLAIR as input modalities. The two models in the real and hypercomplex domain yield a very similar Dice score average of about $0.82$ on BraTS19 \cite{Brats2015Menze}, as reported in the bottom part of Tab.~\ref{tab:brats}. This demonstrates the validity of the proposed approach also for finer-scaled tasks. Furthermore, we believe that these results could be improved in future work that specifically targets segmentation tasks.

\subsection{Visualizing multi-view learning}

\begin{figure}[t]
    \centering
    \includegraphics[width=\linewidth]{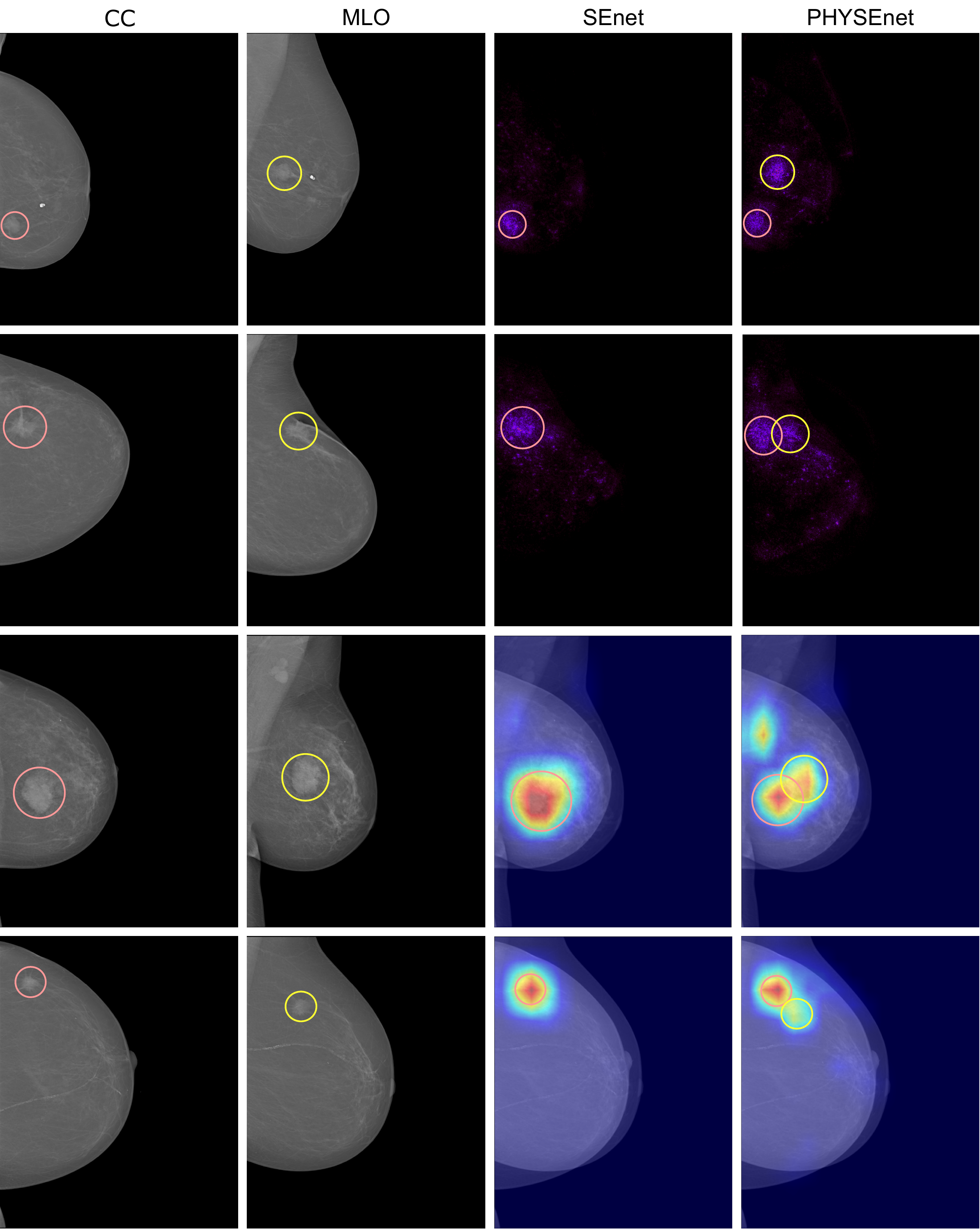}
    \caption{Saliency (top two) and Grad-CAM maps (bottom two) of SEnet and PHYSEnet from different patients. Circles indicate malignant masses, while red and yellow correspond to the CC and MLO view, respectively. Observing the highlighted areas, it is visible that the hypercomplex network effectively models the multi-view information contained in the two views, unlike its real-valued counterpart.}
    \label{fig:saliencies}
\end{figure}

We provide a visualization of how multi-view hypercomplex learning works utilizing simple activation maps in Fig.~\ref{fig:activations} and established explainability techniques, i.e., Grad-CAM \cite{selvaraju2017gradcam} and saliency maps \cite{simonyan2013saliency} in Fig.~\ref{fig:saliencies}. From these visualizations, it is clear how the hypercomplex model learns from both views as the most important areas of the explanations correspond to the ROI in the two views. On the contrary, the real-valued network seems to overfit on the CC view, which is indeed a typical behavior encountered in the literature \cite{Wu2020MultViews}. Moreover, these visualizations serve as a form of weakly supervised detection, highlighting regions that contribute to the model’s classification decisions.
\section{Conclusions}
\label{sec:conc}

In this paper, we have introduced an innovative approach for breast cancer classification, i.e., multi-view hypercomplex learning. Thanks to hypercomplex algebra properties, the proposed hypercomplex models learn from both global (intra-view) and local (inter-view) relations, effectively leveraging the multi-view information of mammographic views. We have demonstrated the effectiveness of our approach both quantitatively and qualitatively across various scenarios, including mammography exams, other radiographic imaging techniques involving different diseases, and a finer-scaled task. Finer-scaled tasks, including segmentation, weakly or fully supervised lesion detection are an interesting direction to be explored more deeply in extensions of this work.

Despite these promising results, there are still a few limitations to address. A current challenge lies in the lack of large-scale pretrained hypercomplex models, which makes fine-tuning less accessible and often requires pretraining from scratch. To help mitigate this, we release all pretrained weights and models discussed in this paper\footnote{\url{https://github.com/ispamm/PHBreast}}. Additionally, while well-known explainability techniques such as Grad-CAM are sufficient for qualitative assessment in this study, the interpretability of hypercomplex operations remains an open research question. Some early efforts toward dedicated explainability methods are beginning to emerge \cite{lopez2024towards}, and we see this as an important direction for future work, particularly to support clinical adoption.

\balance
\bibliographystyle{IEEEtran}
\bibliography{HBCS.bib}

\begin{thebibliography}{10}
\providecommand{\url}[1]{#1}
\csname url@samestyle\endcsname
\providecommand{\newblock}{\relax}
\providecommand{\bibinfo}[2]{#2}
\providecommand{\BIBentrySTDinterwordspacing}{\spaceskip=0pt\relax}
\providecommand{\BIBentryALTinterwordstretchfactor}{4}
\providecommand{\BIBentryALTinterwordspacing}{\spaceskip=\fontdimen2\font plus
\BIBentryALTinterwordstretchfactor\fontdimen3\font minus
  \fontdimen4\font\relax}
\providecommand{\BIBforeignlanguage}[2]{{%
\expandafter\ifx\csname l@#1\endcsname\relax
\typeout{** WARNING: IEEEtran.bst: No hyphenation pattern has been}%
\typeout{** loaded for the language `#1'. Using the pattern for}%
\typeout{** the default language instead.}%
\else
\language=\csname l@#1\endcsname
\fi
#2}}
\providecommand{\BIBdecl}{\relax}
\BIBdecl

\bibitem{Siegel2022Cancer}
R.~L. Siegel, K.~D. Miller, H.~E. Fuchs, and A.~Jemal, ``Cancer statistics,
  2022,'' \emph{CA: A Cancer Journal for Clinicians}, vol.~72, no.~1, pp.
  7--33, 2022.

\bibitem{MOREIRA2012236}
I.~C. Moreira \emph{et~al.}, ``{IN}breast: Toward a full-field digital
  mammographic database,'' \emph{Academic Radiology}, vol.~19, no.~2, pp.
  236--248, 2012.

\bibitem{Misra2010Screening}
S.~Misra, N.~L. Solomon, F.~L. Moffat, and L.~G. Koniaris, ``Screening criteria
  for breast cancer,'' \emph{Adv. Surg.}, vol.~44, pp. 87--100, 2010.

\bibitem{Gur2009DBT}
D.~Gur \emph{et~al.}, ``Digital breast tomosynthesis: Observer performance
  study,'' \emph{American Journal of Roentgenology}, vol. 193, no.~2, pp.
  586--591, 2009.

\bibitem{Liu2021Act}
Y.~Liu, F.~Zhang, C.~Chen, S.~Wang, Y.~Wang, and Y.~Yu, ``Act like a
  radiologist: Towards reliable multi-view correspondence reasoning for
  mammogram mass detection,'' \emph{IEEE Trans. Pattern Anal. Mach. Intell.},
  no.~01, pp. 1--1, 2021.

\bibitem{zhou2025adaptive}
M.~Zhou, G.~Li, C.~Shang, S.~Jin, J.~Lin, L.~Shen, N.~Naik, J.~Peng, and
  Q.~Shen, ``Adaptive fuzzy transformation for abnormal breast mass
  detection,'' \emph{Knowledge-Based Systems}, p. 114232, 2025.

\bibitem{alhussen2025xai}
A.~Alhussen, M.~A. Haq, A.~A. Khan, R.~K. Mahendran, and S.~Kadry,
  ``{XAI-RACapsNet: Relevance} aware capsule network-based breast cancer
  detection using mammography images via explainability {O}-net {ROI}
  segmentation,'' \emph{Expert Systems with Applications}, vol. 261, p. 125461,
  2025.

\bibitem{liang2024hradnet}
Y.~Liang, W.~Tang, T.~Wang, W.~W. Ng, S.~Chen, K.~Jiang, X.~Wei, X.~Jiang, and
  Y.~Guo, ``{HR}ad{N}et: A hierarchical radiomics-based network for multicenter
  breast cancer molecular subtypes prediction,'' \emph{{IEEE} Trans. Med.
  Imag.}, vol.~43, no.~3, pp. 1225--1236, 2024.

\bibitem{li2025interpretable}
G.~Li, M.~Zhou, Y.~Fu, N.~Alam, E.~Denton, and R.~Zwiggelaar, ``An
  interpretable cnn-based model for mass classification in mammography,''
  \emph{Knowledge-Based Systems}, vol. 316, p. 113372, 2025.

\bibitem{shen2019Mammo}
L.~Shen, L.~Margolies, J.~Rothstein, E.~Fluder, R.~McBride, and W.~Sieh, ``Deep
  learning to improve breast cancer detection on screening mammography,''
  \emph{Sci. Rep.}, vol.~9, 2019.

\bibitem{mo2023hover}
Y.~Mo \emph{et~al.}, ``Ho{V}er-{T}rans: Anatomy-aware hover-transformer for
  {ROI}-free breast cancer diagnosis in ultrasound images,'' \emph{{IEEE}
  Trans. Med. Imag.}, vol.~42, no.~6, pp. 1696--1706, 2023.

\bibitem{iqbal2023bts}
A.~Iqbal and M.~Sharif, ``{BTS-ST}: {S}win transformer network for segmentation
  and classification of multimodality breast cancer images,''
  \emph{Knowledge-Based Systems}, vol. 267, p. 110393, 2023.

\bibitem{Wu2021ReducingFB}
N.~Wu \emph{et~al.}, ``Reducing false-positive biopsies using deep neural
  networks that utilize both local and global image context of screening
  mammograms,'' \emph{Journal of Digit. Imag.}, vol.~34, pp. 1414 -- 1423,
  2021.

\bibitem{mahmood2024harnessing}
T.~Mahmood, T.~Saba, A.~Rehman, and F.~S. Alamri, ``Harnessing the power of
  radiomics and deep learning for improved breast cancer diagnosis with
  multiparametric breast mammography,'' \emph{Expert Systems with
  Applications}, vol. 249, p. 123747, 2024.

\bibitem{ren2023ipsilateral}
Y.~Ren, X.~Liu, J.~Ge, Z.~Liang, X.~Xu, L.~J. Grimm, J.~Go, J.~R. Marks, and
  J.~Y. Lo, ``Ipsilateral lesion detection refinement for tomosynthesis,''
  \emph{{IEEE} Trans. Med. Imag.}, vol.~42, no.~10, pp. 3080--3090, 2023.

\bibitem{sreekala2025enhancing}
K.~Sreekala and J.~Sahoo, ``Enhancing mammogram classification using
  explainable conditional self-attention generative adversarial network,''
  \emph{Expert Systems with Applications}, p. 128640, 2025.

\bibitem{wang2022wdccnet}
Y.~Wang, Z.~Wang, Y.~Feng, and L.~Zhang, ``{WDCCN}et: Weighted
  double-classifier constraint neural network for mammographic image
  classification,'' \emph{{IEEE} Trans. Med. Imag.}, vol.~41, no.~3, pp.
  559--570, 2022.

\bibitem{shen2021interpretable}
Y.~Shen, N.~Wu, J.~Phang, J.~Park, K.~Liu, S.~Tyagi, L.~Heacock, S.~G. Kim,
  L.~Moy, K.~Cho \emph{et~al.}, ``An interpretable classifier for
  high-resolution breast cancer screening images utilizing weakly supervised
  localization,'' \emph{Med. Imag. Anal.}, vol.~68, p. 101908, 2021.

\bibitem{lopez2024attention}
E.~Lopez, F.~Betello, F.~Carmignani, E.~Grassucci, and D.~Comminiello,
  ``Attention-map augmentation for hypercomplex breast cancer classification,''
  \emph{Pattern Recognit. Lett.}, vol. 182, pp. 140--146, 2024.

\bibitem{prinzi2024breast}
F.~Prinzi, A.~Orlando, S.~Gaglio, and S.~Vitabile, ``Breast cancer
  classification through multivariate radiomic time series analysis in
  {DCE-MRI} sequences,'' \emph{Expert Systems with Applications}, vol. 249, p.
  123557, 2024.

\bibitem{Wu2020Screen}
N.~Wu, J.~Phang, J.~Park, Y.~Shen \emph{et~al.}, ``Deep neural networks improve
  radiologists' performance in breast cancer screening,'' \emph{IEEE Trans. on
  Med. Imaging}, vol. 39(4), 2020.

\bibitem{Wu2020MultViews}
N.~Wu, S.~Jastrz\k{e}bski, J.~Park, L.~Moy, K.~Cho, and K.~Geras, ``Improving
  the ability of deep neural networks to use information from multiple views in
  breast cancer screening,'' in \emph{Med. Imag. with Deep Learn.}, vol.
  121.\hskip 1em plus 0.5em minus 0.4em\relax PMLR, 2020, pp. 827--842.

\bibitem{Zhang2020NewConv}
C.~Zhang, J.~Zhao, J.~Niu, and D.~Li, ``New convolutional neural network model
  for screening and diagnosis of mammograms,'' \emph{PLoS ONE}, vol. 15(8),
  2020.

\bibitem{Kyono2018MAMMOAD}
T.~Kyono, F.~J. Gilbert, and M.~van~der Schaar, ``{MAMMO}: A deep learning
  solution for facilitating radiologist-machine collaboration in breast cancer
  diagnosis,'' \emph{arXiv preprint: arXiv:1811.02661}, 2018.

\bibitem{Kyono2021Triage}
------, ``Triage of 2{D} mammographic images using multi-view multi-task
  convolutional neural networks,'' \emph{ACM Trans. Comput. Healthcare},
  vol.~2, no.~3, 2021.

\bibitem{kyono2019multi}
------, ``Multi-view multi-task learning for improving autonomous mammogram
  diagnosis,'' in \emph{Mach. Learn, for Healthcare Conf.}\hskip 1em plus 0.5em
  minus 0.4em\relax PMLR, 2019, pp. 571--591.

\bibitem{Sun2019MVConv}
L.~Sun, J.~Wang, Z.~Hu, Y.~Xu, and Z.~Cui, ``Multi-view convolutional neural
  networks for mammographic image classification,'' \emph{IEEE Access}, vol.~7,
  pp. 126\,273--126\,282, 2019.

\bibitem{YANG2021MommiNet}
Z.~Yang \emph{et~al.}, ``Mommi{N}et-v2: Mammographic multi-view mass
  identification networks,'' \emph{Med. Image Anal.}, vol.~73, p. 102204, 2021.

\bibitem{wu2024local}
W.~Wu, Q.~Rong, and Z.~Lu, ``Local cross-view transformers and global
  representation collaborating for mammogram classification,'' \emph{{IEEE}
  Access}, vol.~12, pp. 74\,596--74\,606, 2024.

\bibitem{han2024deep}
B.~Han, L.~Sun, C.~Li, Z.~Yu, W.~Jiang, W.~Liu, D.~Tao, and B.~Liu, ``Deep
  location soft-embedding-based network with regional scoring for mammogram
  classification,'' \emph{{IEEE} Trans. Med. Imag.}, 2024.

\bibitem{nakach2024comprehensive}
F.-Z. Nakach, A.~Idri, and E.~Goceri, ``A comprehensive investigation of
  multimodal deep learning fusion strategies for breast cancer
  classification,'' \emph{Artificial Intelligence Review}, vol.~57, no.~12, p.
  327, 2024.

\bibitem{GaudetIJCNN2018}
C.~Gaudet and A.~Maida, ``Deep quaternion networks,'' in \emph{{IEEE} Int.
  Joint Conf. on Neural Netw. ({IJCNN})}, Rio de Janeiro, Brazil, Jul. 2018.

\bibitem{sigillo2025quaternion}
L.~Sigillo, C.~Bianchi, A.~Uncini, and D.~Comminiello, ``Quaternion
  wavelet-conditioned diffusion models for image super-resolution,'' \emph{Int.
  Joint Conf. on Neural Netw. ({IJCNN})}, 2025.

\bibitem{ParcolletAIR2019}
T.~Parcollet, M.~Morchid, and G.~Linar\`es, ``A survey of quaternion neural
  networks,'' \emph{Artif. Intell. Rev.}, Aug. 2019.

\bibitem{ParcolletICLR2019}
T.~Parcollet \emph{et~al.}, ``Quaternion recurrent neural networks,'' in
  \emph{Int. Conf. on Learn. Representations ({ICLR})}, 2019, pp. 1--19.

\bibitem{ComminielloICASSP2019a}
D.~Comminiello, M.~Lella, S.~Scardapane, and A.~Uncini, ``Quaternion
  convolutional neural networks for detection and localization of 3{D} sound
  events,'' in \emph{{IEEE} Int. Conf. on Acoust., Speech and Signal Process.
  ({ICASSP})}, Brighton, UK, May 2019, pp. 8533--8537.

\bibitem{Brignone2022ISCAS}
C.~Brignone, G.~Mancini, E.~Grassucci, A.~Uncini, and D.~Comminiello,
  ``Efficient sound event localization and detection in the quaternion
  domain,'' \emph{IEEE Trans. on Circuits and Syst. II: Express Brief},
  vol.~69, no.~5, pp. 2453--2457, 2022.

\bibitem{grassucci2023grouse}
E.~Grassucci, L.~Sigillo, A.~Uncini, and D.~Comminiello, ``{GROUSE}: A task and
  model agnostic wavelet-driven framework for medical imaging,'' \emph{IEEE
  Signal Process. Letters}, vol.~30, pp. 1397--1401, 2023.

\bibitem{comminiello2024demystifying}
D.~Comminiello, E.~Grassucci, D.~P. Mandic, and A.~Uncini, ``Demystifying the
  hypercomplex: Inductive biases in hypercomplex deep learning,'' \emph{{IEEE}
  Signal Process. Mag.}, 2024.

\bibitem{Zhang2021PHM}
A.~Zhang, Y.~Tay, S.~Zhang, A.~Chan, A.~T. Luu, S.~C. Hui, and J.~Fu, ``Beyond
  fully-connected layers with quaternions: Parameterization of hypercomplex
  multiplications with $1/n$ parameters,'' \emph{Int. Conf. on Mach. Learn.
  ({ICML})}, 2021.

\bibitem{grassucci2021lightweight}
E.~Grassucci, A.~Zhang, and D.~Comminiello, ``{PHNN}s: Lightweight neural
  networks via parameterized hypercomplex convolutions,'' \emph{IEEE Trans. on
  Neural Netw. and Learn. Syst.}, pp. 1--13, 2022.

\bibitem{ParcolletICASSP2019a}
T.~Parcollet, M.~Morchid, and G.~Linar\`es, ``Quaternion convolutional neural
  networks for heterogeneous image processing,'' in \emph{{IEEE} Int. Conf. on
  Acoust., Speech and Signal Process. ({ICASSP})}, Brighton, UK, May 2019, pp.
  8514--8518.

\bibitem{Lee2017CBISdataset}
R.~Lee, F.~Gimenez, A.~Hoogi, M.~Kanae, M.~Gorovoy, and D.~Rubin, ``A curated
  mammography data set for use in computer-aided detection and diagnosis
  research,'' \emph{Sci. Data}, vol.~4, 2017.

\bibitem{Chexpert2019Irvin}
J.~Irvin \emph{et~al.}, ``Che{X}pert: A large chest radiograph dataset with
  uncertainty labels and expert comparison,'' in \emph{{AAAI} Conf. on Artif.
  Intell.}, vol.~33, no.~01, 2019, pp. 590--597.

\bibitem{Brats2015Menze}
B.~H. Menze \emph{et~al.}, ``The multimodal brain tumor image segmentation
  benchmark ({BRATS}),'' \emph{{IEEE} Trans. Med. Imag.}, no.~10, pp.
  1993--2024, 2015.

\bibitem{brats2017bakas}
S.~Bakas \emph{et~al.}, ``Advancing the cancer genome atlas glioma {MRI}
  collections with expert segmentation labels and radiomic features,''
  \emph{Sci. data}, vol.~4, no.~1, pp. 1--13, 2017.

\bibitem{thakur2024systematic}
N.~Thakur, P.~Kumar, and A.~Kumar, ``A systematic review of machine and deep
  learning techniques for the identification and classification of breast
  cancer through medical image modalities,'' \emph{Multimedia Tools and
  Applications}, vol.~83, no.~12, pp. 35\,849--35\,942, 2024.

\bibitem{nassif2022breast}
A.~B. Nassif, M.~A. Talib, Q.~Nasir, Y.~Afadar, and O.~Elgendy, ``Breast cancer
  detection using artificial intelligence techniques: A systematic literature
  review,'' \emph{Artificial intelligence in medicine}, vol. 127, p. 102276,
  2022.

\bibitem{liu2020cross}
Y.~Liu, F.~Zhang, Q.~Zhang, S.~Wang, Y.~Wang, and Y.~Yu, ``Cross-view
  correspondence reasoning based on bipartite graph convolutional network for
  mammogram mass detection,'' in \emph{{IEEE/CVF} Conf. on Comput. Vis. and
  Pattern Recognit. ({CVPR})}, 2020, pp. 3812--3822.

\bibitem{zhao2020cross}
X.~Zhao, L.~Yu, and X.~Wang, ``Cross-view attention network for breast cancer
  screening from multi-view mammograms,'' in \emph{{IEEE} Int. Conf. on
  Acoust., Speech and Signal Process. ({ICASSP})}.\hskip 1em plus 0.5em minus
  0.4em\relax IEEE, 2020, pp. 1050--1054.

\bibitem{chen2022multi}
Y.~Chen, H.~Wang, C.~Wang, Y.~Tian, F.~Liu, Y.~Liu, M.~Elliott, D.~J. McCarthy,
  H.~Frazer, and G.~Carneiro, ``Multi-view local co-occurrence and global
  consistency learning improve mammogram classification generalisation,'' in
  \emph{Int. Conf. on Med. Imag. Comput. and Comput.-Assisted Intervention
  (MICCAI)}.\hskip 1em plus 0.5em minus 0.4em\relax Springer, 2022, pp. 3--13.

\bibitem{van2021multi}
G.~Van~Tulder, Y.~Tong, and E.~Marchiori, ``Multi-view analysis of unregistered
  medical images using cross-view transformers,'' in \emph{Int. Conf. on Med.
  Imag. Comput. and Comput.-Assisted Intervention (MICCAI)}.\hskip 1em plus
  0.5em minus 0.4em\relax Springer, 2021, pp. 104--113.

\bibitem{sun2022transformer}
Z.~Sun, H.~Jiang, L.~Ma, Z.~Yu, and H.~Xu, ``Transformer based multi-view
  network for mammographic image classification,'' in \emph{Int. Conf. on Med.
  Imag. Comput. and Comput.-Assisted Intervention (MICCAI)}.\hskip 1em plus
  0.5em minus 0.4em\relax Springer, 2022, pp. 46--54.

\bibitem{sarker2024mvswint}
S.~Sarker, P.~Sarker, G.~Bebis, and A.~Tavakkoli, ``{MV-Swin-T}: Mammogram
  classification with multi-view swin transformer,'' in \emph{IEEE Int. Symp.
  on Biom. Imag. (ISBI)}.\hskip 1em plus 0.5em minus 0.4em\relax IEEE, 2024,
  pp. 1--5.

\bibitem{xu2025understanding}
Y.~Xu, Y.~Shen, C.~Fernandez-Granda, L.~Heacock, and K.~J. Geras,
  ``Understanding differences in applying {DETR} to natural and medical
  images,'' \emph{Machine Learning for Biomedical Imaging}, vol.~3, pp.
  152--170, 2025.

\bibitem{grassucci2022hypercomplex}
E.~Grassucci, L.~Sigillo, A.~Uncini, and D.~Comminiello, ``Hypercomplex
  image-to-image translation,'' in \emph{Int. Joint Conf. on Neural Netw.
  ({IJCNN})}.\hskip 1em plus 0.5em minus 0.4em\relax IEEE, 2022, pp. 1--8.

\bibitem{bojesomo2024deep}
A.~Bojesomo, P.~Liatsis, and H.~Al~Marzouqi, ``Deep hypercomplex networks for
  spatiotemporal data processing: Parameter efficiency and superior
  performance,'' \emph{IEEE Signal Process. Mag.}, vol.~41, no.~3, pp.
  101--112, 2024.

\bibitem{singh2024ux}
S.~Singh, U.~Sharma, and V.~Patel, ``{UX-Net} based speech separation scheme
  using parameterized hypercomplex convolutions,'' in \emph{IEEE Int. Conf. on
  Intell. Signal Process. and Effective Comm. Technol. ({INSPECT})}.\hskip 1em
  plus 0.5em minus 0.4em\relax IEEE, 2024, pp. 1--6.

\bibitem{panagos2024visual}
I.~I. Panagos, G.~Sfikas, and C.~Nikou, ``Visual speech recognition using
  compact hypercomplex neural networks,'' \emph{Pattern Recognit. Letters},
  vol. 186, pp. 1--7, 2024.

\bibitem{lopez2023hypercomplex}
E.~Lopez, E.~Chiarantano, E.~Grassucci, and D.~Comminiello, ``Hypercomplex
  multimodal emotion recognition from eeg and peripheral physiological
  signals,'' in \emph{{IEEE} Int. Conf. on Acoust., Speech and Signal Process.
  Workshops ({ICASSPW})}.\hskip 1em plus 0.5em minus 0.4em\relax IEEE, 2023,
  pp. 1--5.

\bibitem{lopez2024hierarchical}
E.~Lopez, A.~Uncini, and D.~Comminiello, ``Hierarchical hypercomplex network
  for multimodal emotion recognition,'' in \emph{IEEE Int. Workshop on Mach.
  Learn. for Signal Process. ({MLSP})}.\hskip 1em plus 0.5em minus 0.4em\relax
  IEEE, 2024, pp. 1--6.

\bibitem{basso2023efficient}
L.~Basso, Z.~Ren, and W.~Nejdl, ``Efficient {ECG}-based atrial fibrillation
  detection via parameterised hypercomplex neural networks,'' in \emph{European
  Signal Process. Conf. ({EUSIPCO})}.\hskip 1em plus 0.5em minus 0.4em\relax
  IEEE, 2023, pp. 1375--1379.

\bibitem{lopez2024phemonet}
E.~Lopez, A.~Uncini, and D.~Comminiello, ``{PHemoNet}: A multimodal network for
  physiological signals,'' in \emph{IEEE Forum on Res. and Technol. for Soc.
  and Industry Innov. ({RTSI})}.\hskip 1em plus 0.5em minus 0.4em\relax IEEE,
  2024, pp. 260--264.

\bibitem{yang2024parameterized}
W.~Yang, S.~Wei, and L.~Zhang, ``Parameterized hypercomplex convolutional
  network for accurate protein backbone torsion angle prediction,''
  \emph{Scientific Reports}, vol.~14, no.~1, p. 27193, 2024.

\bibitem{sigillo2025generalizing}
L.~Sigillo, E.~Grassucci, A.~Uncini, and D.~Comminiello, ``Generalizing medical
  image representations via quaternion wavelet networks,''
  \emph{Neurocomputing}, vol. 638, p. 130195, 2025.

\bibitem{valle2021hypercomplex}
M.~E. Valle and R.~A. Lobo, ``Hypercomplex-valued recurrent correlation neural
  networks,'' \emph{Neurocomputing}, vol. 432, pp. 111--123, 2021.

\bibitem{GrassucciQGAN2021}
E.~Grassucci, E.~Cicero, and D.~Comminiello, ``Quaternion generative
  adversarial networks,'' in \emph{Generative Adversarial Learning:
  Architectures and Applications}, R.~Razavi-Far, A.~Ruiz-Garcia, V.~Palade,
  and J.~Schmidhuber, Eds.\hskip 1em plus 0.5em minus 0.4em\relax Cham:
  Springer International Publishing, 2022, pp. 57--86.

\bibitem{brats2018bakas}
S.~Bakas \emph{et~al.}, ``Identifying the best machine learning algorithms for
  brain tumor segmentation, progression assessment, and overall survival
  prediction in the {BRATS} challenge,'' \emph{arXiv preprint
  arXiv:1811.02629}, 2018.

\bibitem{Post-hocOSCarneiro2021}
R.~Hermoza, G.~Maicas, J.~C. Nascimento, and G.~Carneiro, ``Post-hoc overall
  survival time prediction from brain {MRI},'' in \emph{{IEEE} Int. Symp. on
  Biom. Imag. ({ISBI})}, 2021, pp. 1476--1480.

\bibitem{Resnet2016}
K.~He, X.~Zhang, S.~Ren, and J.~Sun, ``Deep residual learning for image
  recognition,'' in \emph{{IEEE/CVF} Conf. on Comput. Vis. and Pattern
  Recognit. ({CVPR})}, 2016, pp. 770--778.

\bibitem{petrini2022breast}
D.~G. Petrini \emph{et~al.}, ``Breast cancer diagnosis in two-view mammography
  using end-to-end trained {E}fficient{N}et-based convolutional network,''
  \emph{IEEE Access}, vol.~10, pp. 77\,723--77\,731, 2022.

\bibitem{Rubin2018LargeSA}
J.~Rubin, D.~Sanghavi, C.~Zhao, K.~Lee, A.~Qadir, and M.~Xu-Wilson, ``Large
  scale automated reading of frontal and lateral chest {X}-rays using dual
  convolutional neural networks,'' \emph{arXiv preprint: arXiv:1804.07839},
  2018.

\bibitem{deepauc2021yuan}
Z.~Yuan, Y.~Yan, M.~Sonka, and T.~Yang, ``Large-scale robust deep {AUC}
  maximization: A new surrogate loss and empirical studies on medical image
  classification,'' in \emph{{IEEE} Int. Conf. on Comput. Vis. ({ICCV})}, 2021,
  pp. 3040--3049.

\bibitem{UNetRonneberger2015}
O.~Ronneberger, P.~Fischer, and T.~Brox, ``U-{N}et: Convolutional networks for
  biomedical image segmentation,'' in \emph{Med. Image Comput. and
  Comput.-Assisted Intervention ({MICCAI})}.\hskip 1em plus 0.5em minus
  0.4em\relax Springer International Publishing, 2015, pp. 234--241.

\bibitem{selvaraju2017gradcam}
R.~R. Selvaraju, M.~Cogswell, A.~Das, R.~Vedantam, D.~Parikh, and D.~Batra,
  ``Grad-{CAM}: Visual explanations from deep networks via gradient-based
  localization,'' in \emph{{IEEE} Int. Conf. on Comput. Vis. ({ICCV})}, 2017,
  pp. 618--626.

\bibitem{simonyan2013saliency}
K.~Simonyan, A.~Vedaldi, and A.~Zisserman, ``Deep inside convolutional
  networks: Visualising image classification models and saliency maps,''
  \emph{arXiv preprint arXiv:1312.6034}, 2013.

\bibitem{lopez2024towards}
E.~Lopez, E.~Grassucci, D.~Capriotti, and D.~Comminiello, ``Towards explaining
  hypercomplex neural networks,'' in \emph{{IEEE} Int. Joint Conf. on Neural
  Netw. ({IJCNN})}.\hskip 1em plus 0.5em minus 0.4em\relax IEEE, 2024, pp.
  1--8.

\end{thebibliography}

\end{document}